%% file: main.tex
\documentclass[runningheads]{llncs}
\usepackage[T1]{fontenc}
\usepackage{graphicx}
\usepackage{booktabs}
\usepackage[misc]{ifsym}
\newcommand{\corr}{(\Letter)}

\usepackage{amsmath}
\usepackage[colorlinks=true, allcolors=blue]{hyperref}
\usepackage{multirow}
\usepackage{tikz}
\usepackage{pgfplots}
\pgfplotsset{compat=1.18}
\usepackage[caption=false]{subfig}

\usepackage{bm}
\usepackage{amssymb}
\usepackage{amsfonts}
\usepackage{placeins}
\usepackage{makecell}
\newcommand{\imgcell}[2]{%
\begin{tikzpicture}
  \sbox0{\includegraphics[width=\linewidth]{#1}}%

  \node[anchor=south west, inner sep=0] (img) at (0,0) {\usebox0};

  \node[
    anchor=north west,
    inner sep=0pt,
    fill=white,
    fill opacity=1,
    minimum width=\wd0,
    minimum height=0.45cm 
  ] (topbar) at (img.north west) {};

  \node[
    anchor=north,
    at=(img.north),
    inner sep=2pt,
    fill=white,
    fill opacity=0.75,
    text opacity=1,
    rounded corners=1pt,
    font=\scriptsize
  ] {#2};
\end{tikzpicture}%
}

\newcounter{subfig}[figure]

\begin{document}

\title{Measuring Monosemanticity in Sparse Autoencoders via Latent Activation Coherence}

\titlerunning{Preprint: Monosemanticity via Sparse Autoencoders Latent Coherence}

\author{Katarzyna Filus\inst{1} \corr \and Sebastian Pokuci\'nski\inst{2}}

\authorrunning{K. Filus and S. Pokuci\'nski}

\institute{Institute of Theoretical and Applied Informatics, Polish Academy of Sciences, Bałtycka 5, 44-100 Gliwice, Poland \and
Department of Applied Informatics, Silesian University of Technology, Akademicka 16, 44-100 Gliwice, Poland}

\maketitle              

\begin{abstract}
Within Explainable Artificial Intelligence, mechanistic interpretability uses Sparse Autoencoders (SAEs) to extract more interpretable features from neural representations. However, assessing their monosemanticity, and thus explanation quality, remains challenging. Existing metrics require external concept labels or depend on pretrained embedding models, making them sensitive to encoder's geometry. We introduce the Tversky Monosemanticity Score (TMS), a label-free metric that operationalizes monosemanticity as activation-set coherence of binarized SAE latents, and does not require external embedding encoders. We evaluate TMS on SAEs trained on features from pretrained vision and vision-language models (DINOv3, CLIP, BLIP2), two common SAE regimes (TopK, BatchTopK), multiple sparsity levels, and expansion factors. Our results show that TMS is less affected by encoder anisotropy than its embedding-based alternative, while remaining aligned with established monosemanticity indicators. TMS also reveals distinct SAE training dynamics across base models. Moreover, under encoder anisotropy, TMS provides a stronger indication of probe-based concept deletion effectiveness, while being competitive otherwise.

\end{abstract}

\section{Introduction}

Within Explainable Artificial Intelligence (XAI), mechanistic interpretability aims to improve the transparency of deep networks by analyzing their internal representations. A crucial challenge is feature entanglement, where individual neurons encode multiple unrelated concepts, making representations difficult to interpret \cite{bereska2024mechanistic}. Sparse autoencoders (SAEs) emerged as a key technique in this area, encouraging more disentangled features by projecting activations into higher-dimensional sparse latent spaces \cite{gao2024topk,zaigrajew2025interpreting}.
In this context, monosemanticity plays a key role \cite{bereska2024mechanistic}, describing whether a neuron corresponds to a single coherent concept and reflects reduced feature entanglement. Beyond serving as interpretability indicator, monosemanticity holds functional importance and aids in targeted manipulations of model behavior \cite{harle2025measuring}.

Besides its acknowledged importance, currently there is no consensus on how monosemanticity of SAE neurons should be defined and measured. Existing approaches are grounded in supervised concept-predictive evaluation \cite{harle2025measuring,fereidouni2025evaluating} or similarity among stimuli that activate a given neuron \cite{pach2025sparse}. While these methods provide useful proxies, they typically rely on external resources such as annotated concept labels \cite{harle2025measuring,fereidouni2025evaluating}, auxiliary classifiers \cite{harle2025measuring}, or pretrained embedding models \cite{pach2025sparse}. While methods based on embedding models eliminate the need for concept labels \cite{pach2025sparse}, their estimates still largely depend on the representational geometry of external encoders, and factors such as anisotropy. As a consequence, monosemanticity estimates may vary substantially across encoders, complicating direct comparisons between models. These observations highlight the need to develop a monosemanticity measure independent of external labels and embedding models.

Motivated by the limitations of existing monosemanticity metrics, we introduce a novel Tversky Monosemantic Score (TMS), a metric that operationalizes monosemanticity as activation-set coherence within sparse autoencoder latent features. TMS is independent of both external concept labels and test-time reference encoders. The proposed formulation uses a set-based, non-geometric similarity measure inspired by symmetric Tversky's similarity model from cognitive psychology \cite{tversky1977features}. By operating on binary activation patterns rather than continuous embeddings, TMS is less sensitive to factors induced by embedding geometry, including anisotropy, that distort similarity-based measures. 
This independence from external resources simplifies the evaluation and makes it more self-contained. We evaluate TMS on SAEs trained on representations from pretrained vision and vision-language models across different sparsity regimes and expansion factors, using Mini-ImageNet as a natural image benchmark. Our results show that, despite TMS' simplicity, it remains aligned with established monosemanticity indicators, while showing stronger agreement with the MonoSemanticity Score (MS) \cite{pach2025sparse} in low-anisotropy settings (DINOv3). TMS trajectories show that monosemanticity is not static but evolves during training, with clear differences across SAE regimes and base models, pointing to a direction for further investigation in SAE research. TMS as a model-level measure exhibits stronger association with target drop in linear probes trained on SAE activations, compared to the reference measure for anisotropic embedders such as CLIP. We provide our code in an anonymous Zenodo repository for reproducibility~\cite{anonymous_zenodo_repo}.

\section{Related work}

Mechanistic interpretability aims to uncover the internal structures responsible for neural networks behavior \cite{olah2018building}. A key challenge is \emph{polysemanticity}, meaning that individual neurons in a network respond to multiple unrelated concepts, preventing straightforward interpretation. It is often connected to the superposition hypothesis stating that networks encode more features than neurons through overlapping activation directions \cite{huben2023sparse}. Sparse autoencoders (SAEs) are a promising approach to decompose activations into sparse feature dictionaries that are more disentangled and interpretable, capturing more coherent, single-concept representations -- a property commonly referred to as \emph{monosemanticity}. SAE-based analyses have demonstrated monosemantic feature discovery not only in language models, which were the main focus of the initial studies \cite{bricken2023monosemanticity,gao2024topk,rajamanoharan2024jumping}, but also in vision--language or vision settings, where learned features also align with human concepts and can be used for targeted model steering \cite{pach2025sparse,cywinski2025saeuron}.

Available monosemanticity measures include Feature Monosemanticity Score (FMS) \cite{harle2025measuring}, Jensen-Shannon divergence-based score (JSDS) \cite{fereidouni2025evaluating}, and MonoSemanticity score (MS) \cite{pach2025sparse}. FMS uses performance difference for removed or added features in concept classification. JSDS measures how activations separate across concepts with Jensen–Shannon divergence. The primary limitation of JSD and FMS is that they need external concept labels. A recent study introduced the Monosemanticity Score (MS) \cite{pach2025sparse} to avoid this limitation. MS treats a latent neuron as monosemantic if the input features that activate it are mutually similar, by embedding each image with an external encoder and computing a weighted mean cosine similarity between embeddings. While this is an important step toward label-free measures, it remains dependent on the external encoder geometry, and on the encoder choice, which adds variability and complicates cross-model comparisons. Contrastingly, our Tversky Monosemanticity Score is independent of both external concept labels and embedding models. Moreover, while prior studies typically focus on SAEs trained over a single-type modality (e.g. image-text \cite{pach2025sparse}, text \cite{harle2025measuring}), we provide an analysis across different modality types -- purely visual (DINOv3) and multimodal (CLIP, BLIP2), under consistent SAE training. Finally, beyond the predominantly static monosemanticity evaluation \cite{pach2025sparse,harle2025measuring,fereidouni2025evaluating}, we complement our analysis with a dynamic tracking of monosemanticity training evolution, offering a broader view on SAE interpretability.

\section{Tversky Monosemanticity Score}

We define our novel metric -- \textbf{Tversky Monosemantic Score (TMS)} -- following the feature-based similarity model of Tversky~\cite{tversky1977features} from cognitive psychology to operationalize monosemanticity.  $\mathcal{A} = \{a,b,c,\dots\}$ denotes the domain of stimuli (images), which are fed to the deep learning model under interpretation. 
In Tversky’s theory, each object is represented as a set of binary features, and similarity between two objects depends on their shared and distinctive features. 
For objects $a,b \in \mathcal{A}$ with feature sets 
$A$ and $B$, similarity can be defined as a function
$S(a,b) = F(A \cap B,\, A \setminus B,\, B \setminus A)$,
increasing with the number of shared features and decreasing with the number of distinctive features.

In our case, we interpret a trained Sparse Autoencoder (SAE) as a 
\emph{concept extractor}. Each latent neuron of the trained SAE is treated as a learned \emph{concept proxy} (in the sense of Tversky's feature) and as a detector of recurring activation patterns, rather than a ground-truth semantic concept. Each stimulus (features generated with a model under interpretation for a given image) is described by the set of active latent concepts. 
Let $B_n \subseteq \{1,\dots,K\}$ denote the binary set of active latent neurons for stimulus $n$, obtained via a predefined activation rule -- following recent work showing that binarization of SAE features can improve interpretability and provide an effective alternative representation \cite{gallifant2025sparse,aswal2025llmsymguard}. 
For a latent neuron $k$ and its active set of samples $\mathcal{A}_k = \{\, n \;:\; k \in B_n \,\}$, for which $k$ activated, we measure similarity between pairs 
$n_i, n_j \in \mathcal{A}_k$ using Tversky index:
\begin{equation}
\mathrm{TI}_{\alpha,\beta}(B_{n_i}, B_{n_j})
=
\frac{|B_{n_i} \cap B_{n_j}|}
{|B_{n_i} \cap B_{n_j}| 
+ \alpha |B_{n_i} \setminus B_{n_j}| 
+ \beta |B_{n_j} \setminus B_{n_i}| }.
\end{equation} For the purpose of this paper, we set $\alpha$ and $\beta$ to 1, which reduces the score to Jaccard index, as a parameter-free default. The Tversky framing, however, provides a principled cognitive-science foundation and naturally accommodates future asymmetric extensions \cite{tversky1977features}. 
A latent neuron is considered more \emph{monosemantic} if, on average, stimuli in its active set share more common than distinctive features, which is reflected in high Tversky's within-set similarity:
$s_k = 
\mathbb{E}_{n_i,n_j \in \mathcal{A}_k}
\left[
\mathrm{TI}_{\alpha,\beta}(B_{n_i}, B_{n_j})
\right]$
If the latent neuron's active set is considered empty, i.e., $\mathcal{A}_k = \varnothing$, its monosemanticity score becomes $s_k = 0$. In practice, we compute $\mathrm{TI}_{\alpha,\beta}$ on binarized latent activation vectors. 
Given activations $Z \in \mathbb{R}^{N \times K}$, we obtain a binary matrix $B \in \{0,1\}^{N \times K}$ by thresholding each latent $k$ from the test set at its mean training activation $\mu_k$ (i.e., $B_{n,k}=\mathbb{I}[Z_{n,k}>\mu_k]$) for a parameter-free, neuron-wise and data-derived threshold, following prior work of \cite{aswal2025llmsymguard} using mean-based thresholds computed on SAE training data for its individual neurons. 
For each latent $k$, and its active set $\mathcal{A}_k=\{n:\,B_{n,k}=1\}$, we estimate the within-set similarity by sampling at max $M$ distinct pairs $(n_i,n_j)$ with $n_i\neq n_j$ uniformly at random from $\mathcal{A}_k$, and averaging similarities.

The score reflects internal \emph{conceptual coherence} of stimuli activating a given latent neuron, rather than geometric properties of the embedding models. The intuition for grounding our operationalization of monosemanticity in latent coherence is that a latent encoding a relatively selective concept may tend to activate on stimuli sharing not only that concept, but also its contextual structure. For example, a `bird wings' latent may often co-occur with `beaks' and `feathers', giving consistent activation patterns, whereas polysemantic latents would produce more heterogeneous ones.

\section{Experimental Setup}

We consider pretrained models from the HuggingFace repository\footnote{\url{https://huggingface.co/}}: a purely visual \textbf{DINO-v3} with ConvNeXt backbone (\textit{facebook/dinov3-convnext-tiny-pretrain-lvd1689m})\cite{simeoni2025dinov3}, and vision-language \textbf{CLIP} (\textit{openai/clip-vit-base-patch32})\cite{radford2021clip} and \textbf{BLIP2} (\textit{Salesforce/blip2-opt-2.7b})\cite{li2023blip}. We refer to them as \textbf{base models} (when used to extract features for SAE training) and \textbf{encoders} (when used to embed images for MS \cite{pach2025sparse} computation). The experiments are conducted on Mini-ImageNet \cite{bib:vinyals2016matching}, whose natural categories facilitate qualitative inspection of SAE features. We use DINOv3's global average pooled vector, and CLIP's/BLIP's CLS tokens as embeddings. We train sparse autoencoders (SAEs) adopting two common strategies: \emph{TopK} and \emph{BatchTopK} \cite{gao2024topk}, using  \texttt{overcomplete}\footnote{\url{https://github.com/KempnerInstitute/overcomplete}}. We vary the active features number $K \in {10,32,64}$ and the expansion factor $\mathrm{exp} \in {2,4,8}$, yielding diverse SAE configurations. We train all SAEs for 25 epochs using the Mean Squared Error loss and AdamW optimizer with $3 \times 10^{-4}$ learning rate and 512 batch size, while applying the \texttt{overcomplete's} reanimation loss with weight $10^{-2}$. For TMS computation, we set the max number of sampled pairs at \(M=1000\). An auxiliary stability analysis on an example SAE (BatchTopK, DINOv3, \(K=32\), expansion factor \(\times 4\)) showed that TMS remains stable across 10 runs with \(M \in \{50, 100, 250, 500, 1000, 2500, 5000\}\), with values consistent to three decimal places and standard deviation on the order of \(10^{-5}\) at \(M=1000\). The number of active neurons that contribute to TMS after binarization remained substantial in all configurations, with average TMS always computed over at least several hundred neurons. Detailed statistics for all $M$ and results on active neurons are provided in Appendices \ref{app:stab} and \ref{app:active_neurons}. We use MonoSemanticity score (MS) \cite{pach2025sparse} as embedding-based monosemanticity reference. The test environment was a PC with an Intel i9 12900 CPU, RTX 3090 24GB GPU, and 196 GB of DDR5 RAM. Our evaluation has several stages:

\begin{enumerate}
    \item \textbf{Embedding anisotropy analysis.} 
    We quantify the anisotropy of the base models when used as external encoders for embedding-dependent MS score \cite{pach2025sparse}. This motivates the need for external embedding-free metrics.
    \item \textbf{Monosemanticity estimation.}
    For all SAEs, we compute the proposed TMS alongside the reference -- MS \cite{pach2025sparse}. To complement quantitative results, we present example top activating images for neurons for all base model.
    \item \textbf{Metric correlation.}
    We compute the correlation between monosemanticity estimators by measuring per-neuron Spearman correlations with p-values  between TMS and alternative statistics (MS, within-active-set variance with the activation set defined via TMS) for neurons with non-zero TMS.
    \item \textbf{Training dynamics.}
    Beyond static evaluation, we examine how monosemanticity evolves during SAE training by tracking TMS and MS values across epochs and visualizing their trajectories using scatter plots.
    \item \textbf{Concept deletion experiment.}
    We evaluate the functional relevance of SAE features via a concept deletion experiment. For each SAE, we train a linear probe classifier on Mini-ImageNet and identify the most relevant SAE features for a given class using probe weights (max values), ensuring independence from monosemanticity metrics. We then analyze target accuracy degradation as the top-$N$ features are removed and compute correlations between MS/TMS and the accuracy drops across SAE configurations.To train the probes, we use a scikit-learn's linear probe implemented as Standard Scaler and logistic regression (saga solver, $C=1.0$, \texttt{max\_iter}=25).
\end{enumerate}

In Appendices \ref{app:acc_sae} and \ref{app:acc_linear}, we report SAE reconstruction and health metrics ($R^2$, in-out cosine similarity, latent $L_0$, dead neuron ratio) and pre-deletion probe accuracies (on SAE latents and base model features) to show that the SAEs and probes are sufficiently functional, making our monosemanticity and concept-deletion experiments under a shared setup meaningful.

\section{Experimental results}

\paragraph{The impact of anisotropy level on monosemanticty level estimation}

Fig.~\ref{fig:anisotropy_and_ms} shows how external encoder geometry can affect monosemanticity estimation based on pairwise sample similarity. We first quantify anisotropy of the feature-source models (Fig.~\ref{fig:aniso_table}) using mean pairwise cosine similarity between L2-normalized embeddings and the explained variance ratio (EVR) of the first principal component (PC1).
Both statistics indicate that CLIP is significantly more anisotropic
(\(\sigma_{\text{cosine}}=0.52\), \(\mathrm{EVR}=0.07\))
than BLIP2 (\(0.33\), \(0.05\)), while DINOv3 has the most isotropic space
(\(0.07\), \(0.02\)).
Higher \(\sigma_{\text{cosine}}\) and EVR suggest a stronger concentration of embeddings along some direction,
which can create a bias and inflate similarity estimates. This effect is reflected in embedding-based monosemanticity measures such as MS \cite{pach2025sparse}. Fig.~\ref{fig:anisotropy_and_ms} shows MS computed over non-zero values for all pretrained models and configurations for TOPK SAE with $K=10$. Pretrained models are used both as base models and external encoders for MS (Fig.~\ref{fig:aniso_barplot}).
MS exhibits a strong dependence on the encoder: CLIP-based MS is consistently the highest,
BLIP2 -- intermediate, and DINOv3 -- the lowest.
This ordering follows the anisotropy ranking from Fig.~\ref{fig:anisotropy_and_ms},
showing that highly anisotropic encoders inflate embedding-based scores, without necessarily reflecting a corresponding increase in monosemanticity.
These observations motivate creating a monosemanticity score that is not dependent of external encoders.

\begin{figure}[t]
    \centering
    \addtocounter{figure}{1}%
    \begin{minipage}[t]{0.47\linewidth}
        \centering
        \vspace{0pt}
        \refstepcounter{subfig}\label{fig:aniso_table}%
        \begin{tabular}{lrr}
            \toprule
            \textbf{Model} & \multicolumn{1}{c}{\textbf{$\bm{\sigma}_{\text{cosine}}$}} & \multicolumn{1}{c}{\textbf{EVR}} \\
            \midrule
            DINOv3 & 0.07 & 0.02 \\
            CLIP & \textbf{0.52} & \textbf{0.07} \\
            BLIP2 & 0.33 & 0.05 \\
            \bottomrule
        \end{tabular}
        \vspace{0.6em}
        \par\raggedright
        {\small \textbf{(a)} Anisotropy statistics of embedding models: mean cosine similarity ($\sigma_{\text{cosine}}$) and the first principal component's explained variance ratio (EVR). Bold: the highest anisotropy.}
    \end{minipage}
    \hfill
    \begin{minipage}[t]{0.49\linewidth}
        \vspace{0pt}
        \refstepcounter{subfig}\label{fig:aniso_barplot}%
        \includegraphics[width=\linewidth]{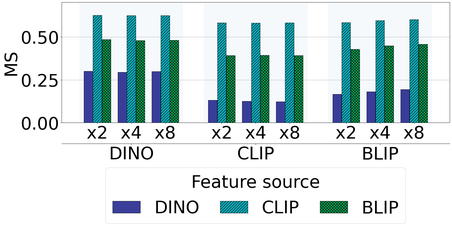}
        \vspace{0.5em}
        {\small \textbf{(b)} Aggregated MS for non-zero neurons (TOPK SAE, K=10).}
    \end{minipage}
    \addtocounter{figure}{-1}%
    \caption{Anisotropy-induced bias in embedding-based MS. (a) External embedding models exhibit different degrees of anisotropy. (b) MonoSemanticity Score (MS) \cite{pach2025sparse} aggregated over non-zero-MS neurons depends strongly on the embedding source---more anisotropic encoders produce inflated MS results across setups.}
    \label{fig:anisotropy_and_ms}
\end{figure}

\begin{figure}[htbp]
  \centering

  \subfloat[TOPK\label{fig:s3}]{%
    \includegraphics[width=0.85\linewidth]{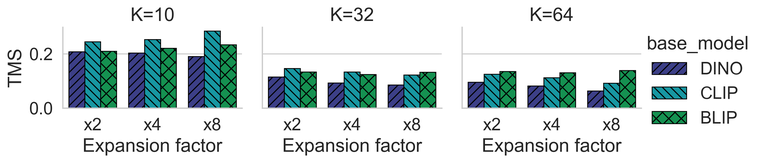}
  }\par\vspace{0.6em}

  \subfloat[BatchTOPK\label{fig:s4}]{%
    \includegraphics[width=0.85\linewidth]{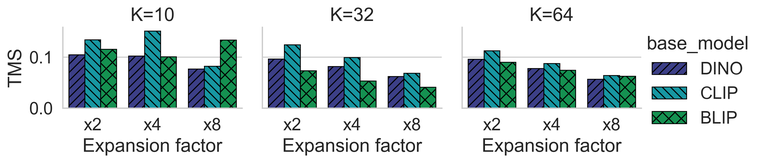}
  }

  \caption{Tversky Monosemantic Score (TMS) results.}
  \label{fig:TMS_barplots}
\end{figure}

\begin{figure}[h]
  \centering

  \subfloat[TOPK\label{fig:s3}]{%
    \includegraphics[width=0.85\linewidth]{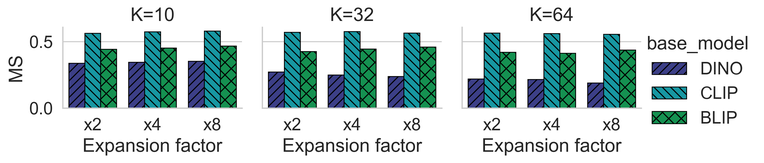}%
  }\\[0.6em]

  \subfloat[BatchTOPK\label{fig:s4}]{%
    \includegraphics[width=0.85\linewidth]{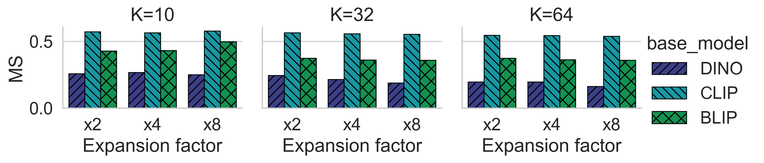}%
  }

  \caption{MonoSemantic (MS) \cite{pach2025sparse} score results.}
  \label{fig:MS_barplots}
\end{figure}

\paragraph{Measuring the monosemanticity level}
In Fig. \ref{fig:TMS_barplots}, we present the barplots of TMS values obtained for all the examined base models and SAEs. Mean is computed for the set of non-zero TMS per-neuron values. We present the reference barplot of MS values in Fig. \ref{fig:MS_barplots} (for the same set of neurons for a fair comparison). In this setup, we use each base model as the encoder for its SAE's MS computation. A clear difference between the two measures emerges. For MS, CLIP dominates with the highest score for all cases. TMS produces a more balanced ordering of models with two examined multimodal models sharing the highest scores -- CLIP (in 13 cases) and BLIP2 (in 5 cases). The overall inter-model gap is notably reduced. When the barplots of TMS and MS are compared in configuration pairs, the agreement between TMS and MS is strongest for DINOv3-based SAEs -- likely due to the smallest score inflation due to anisotropy. It suggests that embedding-based MS is less affected by geometric similarity inflation in the more isotropic regime. In contrast, the divergence between MS and TMS for multimodal models indicates that embedding geometry with higher anisotropy can amplify apparent monosemanticity independently of the feature coherence. Overall, the results suggest that TMS captures a notion of within-feature coherence that is more robust to embedding-space anisotropy, leading to a score less dominated by embedding geometry and more reflective of intrinsic feature coherence. Consequently, differences in monosemanticity across SAEs trained on different feature sources appear less pronounced under TMS than under purely geometry-based measures.

\begin{figure}[h]
    \centering
    \setlength{\tabcolsep}{0pt} 
    \renewcommand{\arraystretch}{0} 

    \begin{tabular}{
        @{}
        p{0.2\textwidth}
        p{0.2\textwidth}
        p{0.2\textwidth}
        p{0.2\textwidth}
        p{0.2\textwidth}
        @{}
    }

    \imgcell{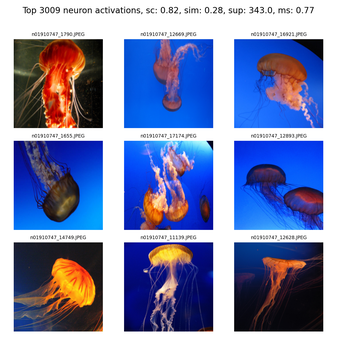}{0.28 (MS = 0.77)} & 
    \imgcell{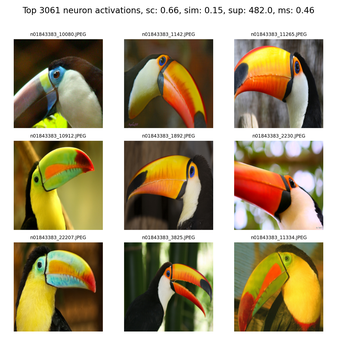}{0.15 (MS = 0.46)} & 
    \imgcell{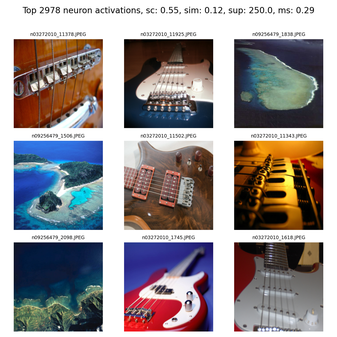}{0.12 (MS = \textbf{0.29})} & 
    \imgcell{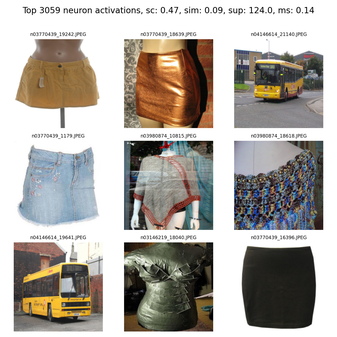}{0.09 (MS = \textbf{0.14})} & 
    \imgcell{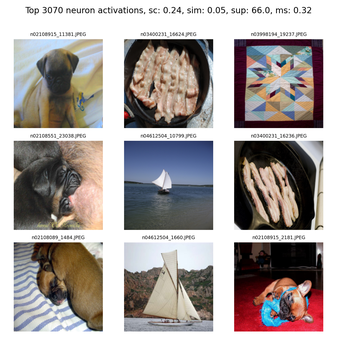}{0.05 (MS = \textbf{0.32})} \\
    
    \imgcell{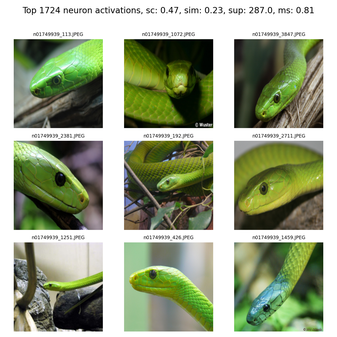}{0.23 (MS = 0.81)} & 
    \imgcell{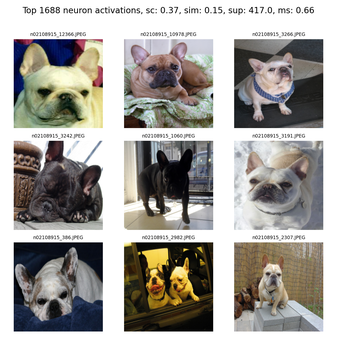}{0.15 (MS = 0.66)} & 
    \imgcell{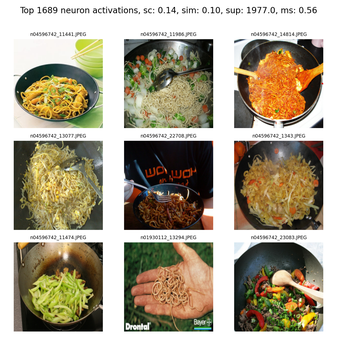}{0.10 (MS = 0.56)} & 
    \imgcell{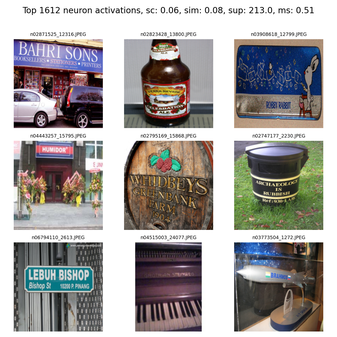}{0.08 (MS = 0.51)} & 
    \imgcell{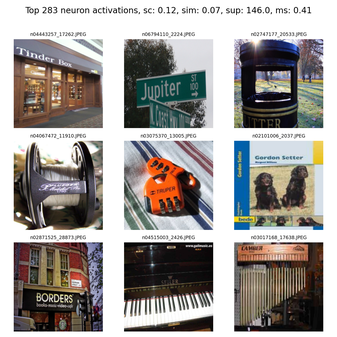}{0.07 (MS = 0.41)} \\
    
    \imgcell{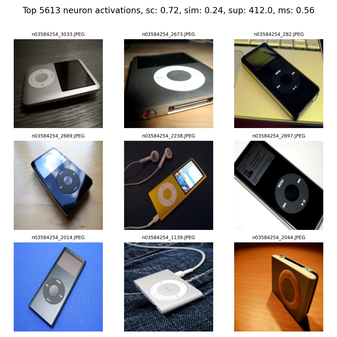}{0.24 (MS = \textbf{0.56})} & 
    \imgcell{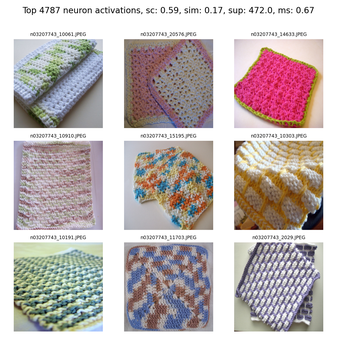}{0.17 (MS = \textbf{0.67})} & 
    \imgcell{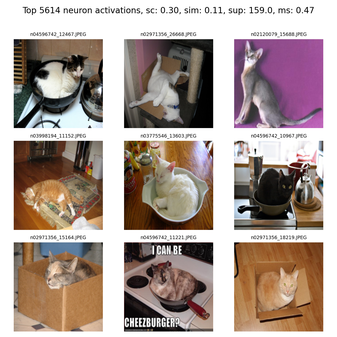}{0.11 (MS = 0.47)} & 
    \imgcell{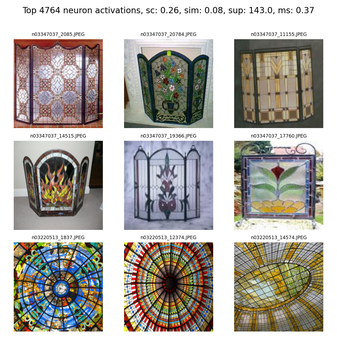}{0.08 (MS = 0.37)} & 
    \imgcell{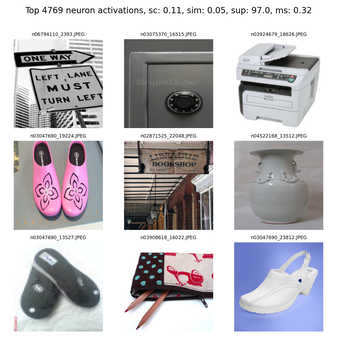}{0.05 (MS = 0.32)} \\

    \end{tabular}
    \caption{Example top 9 activating images for neurons with different TMS. Values decrease from left to right. Top row includes examples for SAE-DINOv3, middle row --- CLIP, and bottom row -- BLIP2. All SAEs are TopK, expansion factor 2 and K = 64. We use bold for cases where ordering is different for TMS and MS.}\label{fig:qual_examples}
\end{figure}

Fig.~\ref{fig:qual_examples} provides qualitative examples of top-activating images for neurons with decreasing TMS values (from left to right) across SAEs trained on DINOv3, CLIP, and BLIP2 features for SAE TopK, $K=64$, expansion factor $2$. A consistent pattern can be observed, in which neurons with high TMS values exhibit visually coherent top activating image sets, sharing a clear semantic/visual motif, whereas lower TMS corresponds to more heterogeneous sets. We  highlight neurons for which the TMS and MS orderings differ. While the overall ranking of neurons is largely consistent across measures, the highlighted cases reveal apparent divergences. 
For example, in the DINOv3 row, neurons with TMS values of $0.12$ and $0.05$ (showing mixtures of islands/musical instruments, and of animals/textures/boats/food, respectively) receive relatively higher MS scores (approximately $0.3$) and an inverse ordering. The latter neuron exhibits a clearly polysemantic activation pattern, and the substantially lower TMS value reflects smaller overlap in its activation sets. For CLIP-based SAEs, neurons with relatively high MS values (e.g. compared to DINOv3) display visually and semantically diverse image sets despite their geometric embedding space proximity. In contrast, TMS assigns lower scores to such neurons, reflecting reduced consistency in their activation behavior. These qualitative observations support the interpretation of TMS as capturing within-feature coherence as proxies for monosemanticity rather than embedding-driven similarity, and provide an intuitive complement to the quantitative results in Figs.~\ref{fig:TMS_barplots} and \ref{fig:MS_barplots}.

\paragraph{Correlation with existing monosemanticity indicators}

To further validate TMS as an indicator of monosemanticity through feature coherence, we analyze its relationship with two neuron-level properties: (i) our main reference -- the MonoSemanticity Score (MS), and (ii) the embedding variance within the activation sets. The second reference serves as an independent structural proxy, as monosemantic neurons are expected to activate on more homogeneous sets of stimuli (expressed with features of base models), which should result in lower variability in their activation sets (based on TMS binarization).

Tab.~\ref{tab:ms_corr} presents the per-neuron correlations between TMS and MS across all SAE variants. MS is computed with a base model used as encoder. Overall, a strong positive relationship is observed, particularly for DINOv3-based SAEs (the least anisotropic encoder), where correlations exceed $0.75$ in most cases (16/18). This indicates that TMS captures a notion of feature monosemanticity largely overlapping with embedding-based measures, despite being computed without reliance on external representations. For CLIP-based SAEs, correlations exhibit greater variability, reflecting the influence of embedding geometry and anisotropy. Nevertheless, moderate to strong correlations are still present in all configurations, showing partial alignment between measures.

In Tab.~\ref{tab:var_corr} we present correlations between TMS and within-activation-set variance of feature embeddings via the base model. Across all models and SAE setups, the correlations are strongly negative, frequently exceeding $-0.9$. This result supports the interpretation that higher TMS values correspond to more coherent activation patterns, aligned with the expected behavior of monosemantic features. The effect is particularly pronounced for BLIP2-based SAEs, where correlations approach $-0.9$ across nearly all configurations. These findings support that TMS reflects both the alignment with existing monosemanticity indicators and coherence of neuron activations. 

\begin{table}[h]
    \centering
    \caption{MS correlation grouped by expansion factor and $K$ for TOPK. All reported correlations are statistically significant ($p$-values $<<$ 0.05).}
    \label{tab:ms_corr}

    \begin{tabular}{llccccccccc}
        \toprule
        SAE &  Model 
        & \multicolumn{3}{c}{x2} 
        & \multicolumn{3}{c}{x4} 
        & \multicolumn{3}{c}{x8} \\
        \cmidrule(lr){3-5} \cmidrule(lr){6-8} \cmidrule(lr){9-11}
        & 
        & K=10 & K=32 & K=64 
        & K=10 & K=32 & K=64 
        & K=10 & K=32 & K=64 \\
        
        \midrule
        TopK 
        & DINOv3 & \textbf{0.88} & \textbf{0.92} & \textbf{0.87} 
               & \textbf{0.87} & \textbf{0.86} & \textbf{0.86} 
               & \textbf{0.86} & \textbf{0.78} & 0.77 \\
        & CLIP & 0.54 & 0.77 & 0.86 
               & 0.53 & 0.75 & 0.83 
               & 0.47 & 0.70 & \textbf{0.78} \\
        & BLIP2 & 0.68 & 0.75 & 0.79 
               & 0.62 & 0.75 & 0.77 
               & 0.64 & 0.75 & 0.75 \\

        \midrule
        BatchTopK 
        & DINOv3 & \textbf{0.87} & \textbf{0.93} & \textbf{0.84} 
               & 0.65          & \textbf{0.89} & 0.89 
               & 0.50          & \textbf{0.78} & 0.85 \\
        & CLIP & 0.35          & 0.60          & 0.90 
               & 0.22          & 0.58          & \textbf{0.90} 
               & 0.27          & 0.61          & \textbf{0.87} \\
        & BLIP2 & 0.81          & 0.79          & 0.82
               & \textbf{0.80} & 0.68          & 0.78 
               & \textbf{0.79} & 0.69          & 0.76 \\

        \bottomrule
    \end{tabular}
\end{table}

\begin{table}[h]
    \centering
    \caption{Variance correlation itself grouped by $K$ and expansion factor for TOPK. All reported correlations are statistically significant ($p$-values $<<$ 0.05).}
    \label{tab:var_corr}

    \begin{tabular}{llccccccccc}
        \toprule
        SAE &  Model 
        & \multicolumn{3}{c}{x2} 
        & \multicolumn{3}{c}{x4} 
        & \multicolumn{3}{c}{x8} \\
        \cmidrule(lr){3-5} \cmidrule(lr){6-8} \cmidrule(lr){9-11}
        & 
        & K=10 & K=32 & K=64 
        & K=10 & K=32 & K=64 
        & K=10 & K=32 & K=64 \\

        \midrule
        TopK 
        & DINOv3 & -0.85 & -0.85 & -0.76 
               & -0.87 & -0.83 & -0.83 
               & -0.84 & -0.80 & -0.82 \\
        & CLIP & -0.81 & -0.90 & -0.88 
               & -0.79 & -0.90 & -0.91 
               & -0.74 & -0.90 & -0.91 \\
        & BLIP2 & \textbf{-0.90} & \textbf{-0.93} & \textbf{-0.92} 
               & \textbf{-0.90} & \textbf{-0.94} & \textbf{-0.94} 
               & \textbf{-0.91} & \textbf{-0.93} & \textbf{-0.93} \\

        \midrule
        BatchTopK 
        & DINOv3 & -0.81 & -0.82 & -0.70 
               & -0.64 & -0.86 & -0.85 
               & -0.49 & -0.81 & -0.88 \\
        & CLIP & -0.45 & -0.71 & -0.92 
               & -0.47 & -0.65 & \textbf{-0.93 }
               & -0.60 & -0.68 & -0.90 \\
        & BLIP2 & \textbf{-0.90} & \textbf{-0.86} & \textbf{-0.88} 
               & \textbf{-0.91} & \textbf{-0.78} & -0.90 
               & \textbf{-0.90} & \textbf{-0.84} & \textbf{-0.89} \\

        \bottomrule
    \end{tabular}
\end{table}

\paragraph{Evolution of monosemanticity during SAE training}

To see how monosemanticty changes over the course of optimization, we analyze the dynamics of MS and TMS throughout SAE training. Figs.~\ref{fig:epoch_evolution_ms_score} and \ref{fig:epoch_evolution_tversky_similarity} report how both measures evolve across epochs for all the base models and SAE configurations. It is visible that both measures have similar trajectory trends for the majority of plots when compared in pairs -- again mostly visible for DINOv3-based SAEs (increase at the beginning, followed by drop and stabilization). Trajectories show that monosemanticity in SAEs trained with different base model features evolves differently (e.g. the trend described for DINOv3 vs the TMS decreasing trend for BLIP2 (without the initial "bump"). TMS shows that monosemanticity of TopK SAEs with smaller K regime ($K=10$)  results in significantly higher scores over training than for higher K values. Contrastingly, for the BatchTopK variant, all SAE configurations converge more closely and obtain lower values. MS exhibits more irregular epoch-to-epoch fluctuations than TMS (e.g. MS for the CLIP-based TopK SAE, in which MS increases quickly in early epochs and then oscillates with relatively high-frequency fluctuations). Despite these differences, both measures tend to display synchronized local fluctuations, indicating that certain training phases induce genuine variability in feature emergence (e.g. for DINOv3-based TopK SAE with k=32/expansion factor 8). These observations suggest that monosemanticity is not static during SAE training but undergoes distinct phases of feature development, while its dynamics are different for SAEs trained with different base model features and sparsity. TMS and MS can capture these phenomena, highlighting the need for such dynamic examinations.

\input{FIGS/scattered/epoch_ms_score}

\input{FIGS/scattered/epoch_tversky_similarity}

\paragraph{Concept deletion effectiveness}

To assess whether model-level TMS monosemanticity translates into effective feature-level interventions for a given SAE, we perform a probe-based concept deletion experiment. For each SAE, we train a linear probe on SAE representations and mini-ImageNet classes. We zero the top-$N$ latent features most relevant to a given class, ranked by the magnitude of the probe weights, treating classes as higher-level semantic targets. The resulting drop in target-class accuracy (target drop) serves as a measure of the functional relevance of SAE features.  Fig.~\ref{fig:deletion_curves_target_per_metric} shows target drop as a function of the number of removed features \(N\). Distinct behaviors emerge across base models. For DINOv3 SAEs, smaller \(K\) and lower expansion factors yield the strongest deletion effects. For CLIP and BLIP2 SAEs, smaller \(K\) also tends to produce stronger target drops at small \(N\) (even \(N=1\)). This is consistent with our earlier TMS results, where these configurations exhibited higher monosemanticity, suggesting that tighter sparsity yields more concentrated features. For CLIP, the \(K=10\) settings show comparable or smaller drops at larger \(N\), likely due to their lower probe accuracies (\(\sim 60\text{--}70\%\)). Similar behavior for BLIP2 BatchTopK SAEs despite high probe accuracies (\(>90\%\)) may indicate greater feature overlap.

To quantify the link between monosemanticity and deletion effectiveness, we compute correlations between SAE-level mean TMS/MS and target drop across all trained SAEs for each \(N\) (Tab.~\ref{tab:corr_drop}). For small \(N\), both metrics show strong positive correlations with target drop for all models, indicating that more monosemantic SAEs enable more effective interventions. For CLIP-based SAEs, correlations become non-significant for \(N=16\) and \(N=32\), likely due in part to lower downstream probe performance. Comparing the two measures, TMS gives stronger correlations for CLIP-based SAEs (where MS is impacted by higher anisotropy), while comparable, but slightly weaker results in less anisotropic settings (DINOv3 and BLIP2 at smaller \(N\)). Overall, both metrics capture functional feature relevance, with TMS offering better predictive value when embedding geometry may distort similarity-based assessments.

\input{FIGS/corr/target_drop_combined_plot}\label{fig:targetdrop}

\begin{table*}[]
    \centering
    \caption{Correlation analysis of TMS/MS-target drop across different models and N values. All correlations computed with corresponding p-values. Best results for each model marked in bold. Non-significant results (p-value > 0.05) in italics.}
    \label{tab:corr_drop}
    \begin{tabular}{|c|c|c|c|c|c|c|c|}
        \hline
        \multirow{2}{*}{Model} & \multirow{2}{*}{N} & \multicolumn{2}{c|}{Target drop - TMS} & \multicolumn{2}{c|}{Target drop - MS}  \\ \cline{3-6}
                               &                    & Correlation & p-value & Correlation & p-value  \\ \hline
        \multirow{6}{*}{DINOv3}  & 1                  & 0.837 & 0.000015 & \textbf{0.878} & 0.000002  \\ \cline{2-6}
                               & 2                  & 0.878 & 0.000002 & \textbf{0.915} & 0.000000 \\ \cline{2-6}
                               & 4                  & 0.876 & 0.000002 & \textbf{0.893} & 0.000001  \\ \cline{2-6}
                               & 8                  & 0.843 & 0.000011 & \textbf{0.905} & 0.000000  \\ \cline{2-6}
                               & 16                 & 0.825 & 0.000026 & \textbf{0.917} & 0.000000  \\ \cline{2-6}
                               & 32                 & 0.868 & 0.000003 & \textbf{0.917} & 0.000000  \\ \hline
        \multirow{6}{*}{CLIP}  & 1                  & \textbf{0.783} & 0.000121 & 0.721 & 0.000728  \\ \cline{2-6}
                               & 2                  & \textbf{0.827} & 0.000023 & 0.707 & 0.001037  \\ \cline{2-6}
                               & 4                  & \textbf{0.864} & 0.000004 & 0.653 & 0.003286  \\ \cline{2-6}
                               & 8                  & \textbf{0.678} & 0.001985 & 0.534 & 0.022588  \\ \cline{2-6}
                               & 16                 & \textit{0.406} & \textit{0.094953} & \textit{0.257} & \textit{0.303309}  \\ \cline{2-6}
                               & 32                 & \textit{0.158} & \textit{0.531483} & \textit{-0.001} & \textit{0.996757}  \\ \hline
        \multirow{6}{*}{BLIP2}  & 1                  & 0.719 & 0.000767 & \textbf{0.818} & 0.000033  \\ \cline{2-6}
                               & 2                  & 0.674 & 0.002166 & \textbf{0.785} & 0.000113  \\ \cline{2-6}
                               & 4                  & 0.622 & 0.005819 & \textbf{0.750} & 0.000335  \\ \cline{2-6}
                               & 8                  & 0.670 & 0.002360 & \textbf{0.771} & 0.000181  \\ \cline{2-6}
                               & 16                 & \textbf{0.697} & 0.001318 & 0.662 & 0.002791  \\ \cline{2-6}
                               & 32                 & \textbf{0.732} & 0.000558 & 0.577 & 0.012195  \\ \hline
    \end{tabular}
\end{table*}

\FloatBarrier

\section{Conclusions}

In this paper, we introduced the Tversky Monosemanticity Score (TMS), a metric that operationalizes monosemanticity as activation-set coherence of binarized SAE latents. Across multiple SAEs trained on DINOv3, CLIP, and BLIP2 features, we showed that TMS is less affected by encoder anisotropy than the reference embedding-based metric, while remaining correlated with established monosemanticity indicators. Moreover, TMS more clearly revealed differences in training dynamics across base models and SAE regimes during optimization. Finally, our probe-based concept deletion study showed that TMS as a model-level metric was more strongly associated with deletion effectiveness under higher encoder anisotropy than the reference while remaining competitive otherwise. By operating on SAE latents, we eliminate the cost connected to external embedding generation of methods relying on external encoders, varying depending on hardware used, which in our experiments was up to app. 30 mins on the BLIP2 test set. Although our method operates on higher-dimensional SAE representations, its similarity computation is based on sparse vectors, thresholding and binary operations, which are typically computationally cheaper than the floating-point arithmetic on dense vectors, and allow implementation-level optimization.

We adopted Tversky’s feature-based similarity model because it provides the conceptual basis for representing latent activations and their comparisons through shared and distinctive components, as a coherence-based measure. TMS's correlation with other monosemanticity indicators supports its use as a useful estimation aid, especially for natural-image data, where concepts often co-occur with reasonable contextual structure. Practically, against degenerate co-activations, the presented within-active-set variance can be used as a complementary sanity check. Our future work will focus on extending the proposed parameter-free baseline with the asymmetric Tversky variant, more advanced binarization~\cite{gallifant2025sparse,aswal2025llmsymguard}, and a broader evaluation across tasks. Nevertheless, our current study with multiple base models, SAE regimes, sparsity levels, expansion factors, a natural-image task and from different perspectives, shows the potential of TMS for SAE verification without external concept labels or encoders.

\begin{credits}

\subsubsection{Preprint notice} This is a preprint version. A shorter version of this paper has been accepted for presentation and publication in the post-workshop proceedings of the 8th International Workshop on eXplainable Knowledge Discovery in Data Mining (XKDD~2026), co-located with ECML~PKDD~2026. The appendix is included only in this preprint and is not part of the peer-reviewed proceedings paper.

\subsubsection{\ackname} This research was supported by the~START Scholarship of the Foundation for Polish Science (FNP) for outstanding young scholars, agreement No. START 017.2025, the~Polish Minister of Science and Higher Education scholarship for outstanding young researchers, agreement No. SMN/20/1546/2024, and the Research Funds for Young Scientists at the Silesian University of Technology (agreements No. 02/100/BKM25/0045 and 02/100/BKM26/0055). The authors utilized LLMs for language polishing under human supervision and take fully responsible for the manuscript.

\subsubsection{\discintname}
The authors have no competing interests. 
\end{credits}

%
%
%
\bibliographystyle{splncs04}
\bibliography{references}

\newpage

\appendix

\begin{center}
    {\Large\bfseries Appendices: Measuring Monosemanticity in Sparse Autoencoders via Latent Activation Coherence}
\end{center}


\setcounter{figure}{0}
\counterwithin{figure}{section} 
\setcounter{table}{0}
\counterwithin{table}{section}

\section{Accuracy of SAEs}\label{app:acc_sae}

In Tab.~\ref{table:sae_training_metrics_df}, we provide the standard training metrics obtained for all the SAE setups used in our study after the final epoch 25: $R^2$, $L_0$, input-output Cosine Similarity and dead neurons ratio. $R^2$ and dead neurons ratio are obtained directly via overcomplete pipeline, while we compute $L_0$ and input-output Cosine Similarity on the 10\% split of our test dataset. Overall, the results indicate that the large majority of SAEs trained with our unified procedure are at least relatively functionally strong. In particular, the minimum input-output Cosine Similarity is 0.6636, 20 out of 27 BatchTopK models exceed 0.8, all TopK models exceed 0.8, and the best-performing setups achieve values above 0.9. Similarly, most models obtain $R^2 > 0.7$, while dead neurons are effectively eliminated due to the reanimation loss recommended in the overcomplete library. Therefore, although a configuration-specific strategy to tune the parameters would likely provide us with even stronger reconstruction metrics, the obtained models are sufficiently functional to support a fair comparative study of monosemanticity under a shared setup.

\input{TABS/stats_with_dead_features_tab}

\section{Accuracy of linear probes}\label{app:acc_linear}

In Tab.~\ref{table:sae_accuracy}, we report the  probe accuracy obtained for the concept deletion experiment before the deletion. Specifically, we include the accuracy of probes trained on raw features extracted from the base models (DINOv3, CLIP, and BLIP), as well as on the latent activations of all SAEs considered in our experiments. The results show that, across all data sources, it is possible to train relatively accurate probes in the 100-class classification setting via mini-ImageNet. The best overall performance is obtained for BLIP, for which all accuracies exceed 0.9, and for DINO, where 18/19 cases achieve accuracy above 0.9. Weaker results are observed for CLIP embeddings and the latents of SAEs trained on CLIP representations. However, even in this case, the majority of accuracies remain above 0.85. Overall, although both the SAEs and the probes were trained under a unified procedure rather than being individually tuned for a particular configuration, the obtained results are sufficiently strong to support the experiments regarding the concept deletion experiment.

\input{TABS/accuracy_comparison_tab}
\FloatBarrier

\section{TMS stability under sampling}\label{app:stab}

To examine the stability of the sampling-based TMS computation, we conducted an additional experiment on an example SAE configuration: BatchTopK trained on DINOv3 features with sparsity $K=32$ and expansion factor $\times 4$, using the same training setup as for all other SAEs considered in the paper. Table~\ref{tab:tversky_stability} reports the mean and standard deviation of the estimated mean TMS for different values of samples used as an upper limit on the number of pairs $M$ over 10 runs. The results show that the estimate is rather stable across $M$ values, with the mean remaining unchanged at approximately $0.084$ to three decimal places across all tested values. However, as could be predicted, the variability slightly decreases as the number of sampled pairs increases: for $K=1000$, which is the value used throughout the paper, the standard deviation is already on the order of $10^{-5}$, indicating sufficient stability. Even smaller budgets appear reasonably stable for fast comparisons in our case, while larger values provide even higher consistency across repetitions.

\begin{table}
\caption{Stability of mean TMS for different numbers of upper limit on the number of sampled pairs ($M$).}
\label{tab:tversky_stability}
\centering
\setlength{\tabcolsep}{10pt}
\begin{tabular}{|r|c|c|}
\hline
$M$  & mean & std \\
\hline
50 & 0.084 & $1.88 \times 10^{-4}$ \\
\hline
100 & 0.084 & $1.99 \times 10^{-4}$ \\\hline
250 & 0.084 & $9.30 \times 10^{-5}$ \\\hline
500 & 0.084 & $6.78 \times 10^{-5}$ \\\hline
1000 & 0.084 & $6.14 \times 10^{-5}$ \\\hline
2500 & 0.084 & $2.94 \times 10^{-5}$ \\\hline
5000 & 0.084 & $2.46 \times 10^{-5}$ \\
\hline
\end{tabular}
\end{table}

\section{Active latent neurons after binarization}
\label{app:active_neurons}

To better contextualize the Tversky Monosemanticity Score (TMS), we report the number of latent neurons with non-empty active sets $A_k$ after applying our mean-based binarization rule in Tab.~\ref{tab:active_neurons_topk_batchtopk}. For reference, the original feature dimensionalities of the analyzed base models are $768$ for DINO, $512$ for CLIP, and $1408$ for BLIP, which correspond to latent dimensionalities scaled by the considered expansion factors $\times 2$, $\times 4$, and $\times 8$. The reported counts indicate how many latent neurons are considered active after binarization, therefore, how many neurons effectively participate in TMS computation. These statistics reveal a difference between TopK and BatchTopK SAEs regimes. For BatchTopK, the number of active neurons is typically closer to the full latent dimensionality, indicating the expected broad utilization of the SAE dictionary, which is especially visible for CLIP-based SAEs. In this case, nearly all neurons become active across configurations. DINOv3-based SAEs also exhibit high coverage, while BLIP2-based SAEs, although still broadly utilized under BatchTopK, remain visibly below full saturation, especially for larger expansion factors. In contrast, TopK SAEs often use only a subset of their latent space after binarization, which depends on both $K$ and expansion factor. For all base models, increasing $K$ leads to a visible increase in the number of active neurons, whereas increasing the expansion factor reduces the fraction of neurons that are active. This means that although larger expansion increases the overall latent dimensionality, the effective binarized dictionary size does not grow proportionally under TopK, which is mostly observed for BLIP-based SAEs. Overall, the raw latent dimensionality can differ from the number of neurons effectively contributing to TMS. Active-neuron statistics therefore can help contextualize the reported TMS values by clarifying and confirming sufficiently large neuron sets for model-level aggregation.

\begin{table}
\centering
\caption{Active neurons grouped by $K$ and expansion factor for TOPK.}
\label{tab:active_neurons_topk_batchtopk}
\begin{tabular}{llccccccccc}
        \toprule
        SAE &  Model 
        & \multicolumn{3}{c}{x2} 
        & \multicolumn{3}{c}{x4} 
        & \multicolumn{3}{c}{x8} \\
        \cmidrule(lr){3-5} \cmidrule(lr){6-8} \cmidrule(lr){9-11}
        & 
        & K=10 & K=32 & K=64 
        & K=10 & K=32 & K=64 
        & K=10 & K=32 & K=64 \\

        \midrule
TopK & DINO & 1131 & 1533 & 1535 & 1460 & 2908 & 3043 & 1811 & 4376 & 5732 \\
& CLIP & 569 & 954 & 1021 & 659 & 1606 & 2015 & 677 & 2466 & 3766 \\
& BLIP & 670 & 1639 & 2300 & 633 & 2192 & 3627 & 630 & 2475 & 4200 \\
\midrule
BatchTopK& DINO & 1535 & 1535 & 1536 & 2714 & 3058 & 3070 & 4492 & 5928 & 6095 \\
& CLIP & 1024 & 1024 & 1024 & 2048 & 2047 & 2046 & 4080 & 4084 & 4088 \\
& BLIP & 2644 & 2735 & 2685 & 5071 & 5282 & 5096 & 9104 & 9613 & 9243 \\
\bottomrule
\end{tabular}
\end{table}

\end{document}

%% file: FIGS/scattered/epoch_ms_score.tex
\begin{figure*}[h]
    \centering

    \subfloat{
        \begin{minipage}[t]{0.49\textwidth}
        \centering
        \begin{tikzpicture}[trim axis left, trim axis right]
            \begin{axis}[
                 width=\textwidth,
                 height=0.85\textwidth,
                 xlabel={Epoch},
                 xlabel style={yshift=4pt},
                 xtick={0,5,10,15,20,25},
                 xmin=-1.5, xmax=26.5,
                 ymin=0.1,
                 ymax=0.55,
                 ytick={0.15,0.2,0.25,0.3,0.35,0.4,0.45,0.5,0.55},
                 ylabel={MS},
                 grid=both,
                 grid style={line width=.1pt, dashed, draw=gray!10},
                 major grid style={line width=.2pt,draw=gray!50},
                 label style={font=\scriptsize},
                 tick label style={font=\scriptsize},
                 every axis title/.style={at={(0.5,1.0)}, above, font=\small},
                 title={DINO: TopKSAE},
                 mark size=0.8pt,
                 line width=0.4pt,
                 scaled y ticks=false,
                 yticklabel style={/pgf/number format/fixed zerofill, /pgf/number format/precision=2},
                 legend to name=legendTopK,
                 legend style={legend columns=3, font=\scriptsize, column sep=0.3cm},
                 axis x line*=bottom,
                 axis y line*=left,
                 ]
                \addplot[color=red, mark=square*, solid] coordinates {
                  (1,0.205291)(2,0.435805)(3,0.455567)(4,0.431668)(5,0.413872)(6,0.408038)(7,0.390449)(8,0.388682)(9,0.367388)(10,0.384571)(11,0.358878)(12,0.383062)(13,0.361340)(14,0.372319)(15,0.351188)(16,0.383821)(17,0.349706)(18,0.368616)(19,0.347485)(20,0.365930)(21,0.346096)(22,0.352231)(23,0.348034)(24,0.349853)(25,0.337575)
                };
                \addlegendentry{Exp=2, K=10}
                \addplot[color=blue, mark=square*, solid] coordinates {
                  (1,0.271151)(2,0.490153)(3,0.482530)(4,0.468650)(5,0.446664)(6,0.445320)(7,0.426199)(8,0.424079)(9,0.409614)(10,0.404802)(11,0.408589)(12,0.395660)(13,0.397574)(14,0.389944)(15,0.385131)(16,0.377776)(17,0.386772)(18,0.373359)(19,0.378266)(20,0.369558)(21,0.365097)(22,0.360779)(23,0.356527)(24,0.358594)(25,0.350273)
                };
                \addlegendentry{Exp=4, K=10}
                \addplot[color=green!50!black, mark=square*, solid] coordinates {
                  (1,0.296924)(2,0.496501)(3,0.487633)(4,0.468364)(5,0.464941)(6,0.450514)(7,0.445553)(8,0.435947)(9,0.417095)(10,0.421665)(11,0.406047)(12,0.405862)(13,0.401031)(14,0.395076)(15,0.389671)(16,0.387911)(17,0.381028)(18,0.381650)(19,0.370215)(20,0.373185)(21,0.365498)(22,0.369031)(23,0.354857)(24,0.360451)(25,0.349895)
                };
                \addlegendentry{Exp=8, K=10}
                \addplot[color=red, mark=triangle*, solid] coordinates {
                  (1,0.157158)(2,0.275954)(3,0.302526)(4,0.331972)(5,0.277535)(6,0.306595)(7,0.276740)(8,0.283803)(9,0.267823)(10,0.275849)(11,0.265826)(12,0.274172)(13,0.267276)(14,0.276877)(15,0.266325)(16,0.272539)(17,0.269167)(18,0.269977)(19,0.264728)(20,0.275298)(21,0.268585)(22,0.274454)(23,0.267366)(24,0.273529)(25,0.269701)
                };
                \addlegendentry{Exp=2, K=32}
                \addplot[color=blue, mark=triangle*, solid] coordinates {
                  (1,0.171199)(2,0.348717)(3,0.334173)(4,0.381491)(5,0.290913)(6,0.373083)(7,0.277705)(8,0.346769)(9,0.267778)(10,0.323010)(11,0.261488)(12,0.313692)(13,0.257209)(14,0.309152)(15,0.254086)(16,0.296210)(17,0.251282)(18,0.302427)(19,0.248568)(20,0.294459)(21,0.248897)(22,0.285299)(23,0.247234)(24,0.280281)(25,0.250657)
                };
                \addlegendentry{Exp=4, K=32}
                \addplot[color=green!50!black, mark=triangle*, solid] coordinates {
                  (1,0.188423)(2,0.414736)(3,0.349870)(4,0.421437)(5,0.301735)(6,0.403523)(7,0.275146)(8,0.391625)(9,0.249785)(10,0.367770)(11,0.244334)(12,0.357881)(13,0.239816)(14,0.345345)(15,0.242102)(16,0.339877)(17,0.235142)(18,0.327506)(19,0.246147)(20,0.320817)(21,0.246189)(22,0.311113)(23,0.245419)(24,0.297079)(25,0.240215)
                };
                \addlegendentry{Exp=8, K=32}
                \addplot[color=red, mark=diamond*, solid] coordinates {
                  (1,0.134272)(2,0.223709)(3,0.223029)(4,0.245869)(5,0.220894)(6,0.243076)(7,0.219611)(8,0.239708)(9,0.228765)(10,0.235561)(11,0.228345)(12,0.237254)(13,0.223378)(14,0.232720)(15,0.224696)(16,0.229294)(17,0.227474)(18,0.229207)(19,0.224147)(20,0.228082)(21,0.224307)(22,0.222598)(23,0.222635)(24,0.221625)(25,0.216228)
                };
                \addlegendentry{Exp=2, K=64}
                \addplot[color=blue, mark=diamond*, solid] coordinates {
                  (1,0.150728)(2,0.264364)(3,0.236993)(4,0.297076)(5,0.222633)(6,0.288747)(7,0.218524)(8,0.279025)(9,0.220712)(10,0.264317)(11,0.212891)(12,0.260147)(13,0.218093)(14,0.240720)(15,0.218678)(16,0.239048)(17,0.217622)(18,0.234086)(19,0.215635)(20,0.226951)(21,0.221687)(22,0.224876)(23,0.217013)(24,0.224339)(25,0.213249)
                };
                \addlegendentry{Exp=4, K=64}
                \addplot[color=green!50!black, mark=diamond*, solid] coordinates {
                  (1,0.166102)(2,0.322603)(3,0.218363)(4,0.334593)(5,0.220367)(6,0.330039)(7,0.212442)(8,0.321454)(9,0.203486)(10,0.301761)(11,0.205501)(12,0.288992)(13,0.200534)(14,0.271563)(15,0.195370)(16,0.264707)(17,0.190385)(18,0.258475)(19,0.192384)(20,0.249158)(21,0.190737)(22,0.235771)(23,0.192667)(24,0.231321)(25,0.185984)
                };
                \addlegendentry{Exp=8, K=64}
            \end{axis}
        \end{tikzpicture}
    \end{minipage}}\hspace{-4em}%
    \subfloat{
        \begin{minipage}[t]{0.49\textwidth}
        \centering
        \begin{tikzpicture}[trim axis left, trim axis right]
            \begin{axis}[
                 width=\textwidth,
                 height=0.85\textwidth,
                 xlabel={Epoch},
                 xlabel style={yshift=4pt},
                 xtick={0,5,10,15,20,25},
                 xmin=-1.5, xmax=26.5,
                 ymin=0.1,
                 ymax=0.55,
                 ytick={0.15,0.2,0.25,0.3,0.35,0.4,0.45,0.5,0.55},
                 ylabel={},
                 grid=both,
                 grid style={line width=.1pt, dashed, draw=gray!10},
                 major grid style={line width=.2pt,draw=gray!50},
                 label style={font=\scriptsize},
                 tick label style={font=\scriptsize},
                 every axis title/.style={at={(0.5,1.0)}, above, font=\small},
                 title={DINO: BatchTopKSAE},
                 mark size=0.8pt,
                 line width=0.4pt,
                 scaled y ticks=false,
                 yticklabel style={/pgf/number format/fixed zerofill, /pgf/number format/precision=2},
                 legend to name=legendBatch,
                 legend style={legend columns=3, font=\scriptsize, column sep=0.3cm},
                 axis x line*=bottom,
                 axis y line*=right,
                 ]
                \addplot[color=red, mark=square*, solid] coordinates {
                  (1,0.203865)(2,0.380082)(3,0.437954)(4,0.378864)(5,0.378369)(6,0.343244)(7,0.344279)(8,0.302460)(9,0.307379)(10,0.284569)(11,0.301607)(12,0.278505)(13,0.284910)(14,0.271072)(15,0.276377)(16,0.271010)(17,0.271746)(18,0.256129)(19,0.267386)(20,0.254398)(21,0.263743)(22,0.247655)(23,0.260206)(24,0.251053)(25,0.256529)
                };
                \addplot[color=blue, mark=square*, solid] coordinates {
                  (1,0.254102)(2,0.438439)(3,0.490665)(4,0.419494)(5,0.462055)(6,0.406971)(7,0.421759)(8,0.347085)(9,0.376602)(10,0.310070)(11,0.341461)(12,0.291663)(13,0.334551)(14,0.272364)(15,0.290363)(16,0.261059)(17,0.289693)(18,0.258830)(19,0.276643)(20,0.251637)(21,0.273917)(22,0.244109)(23,0.269865)(24,0.242100)(25,0.268544)
                };
                \addplot[color=green!50!black, mark=square*, solid] coordinates {
                  (1,0.317597)(2,0.507723)(3,0.530396)(4,0.505887)(5,0.482113)(6,0.453992)(7,0.437849)(8,0.360044)(9,0.349181)(10,0.339002)(11,0.335491)(12,0.301292)(13,0.304235)(14,0.282734)(15,0.294934)(16,0.266808)(17,0.290953)(18,0.258557)(19,0.276136)(20,0.256525)(21,0.262666)(22,0.243909)(23,0.250247)(24,0.249752)(25,0.241974)
                };
                \addplot[color=red, mark=triangle*, solid] coordinates {
                  (1,0.162182)(2,0.198423)(3,0.278384)(4,0.275831)(5,0.283217)(6,0.279751)(7,0.265369)(8,0.261160)(9,0.256295)(10,0.252957)(11,0.256130)(12,0.251837)(13,0.260188)(14,0.246606)(15,0.252544)(16,0.251994)(17,0.251012)(18,0.250747)(19,0.247916)(20,0.252587)(21,0.248308)(22,0.250029)(23,0.242473)(24,0.249968)(25,0.245277)
                };
                \addplot[color=blue, mark=triangle*, solid] coordinates {
                  (1,0.162272)(2,0.214157)(3,0.352000)(4,0.291947)(5,0.306504)(6,0.267027)(7,0.249288)(8,0.237848)(9,0.231673)(10,0.227961)(11,0.227188)(12,0.223188)(13,0.226300)(14,0.227590)(15,0.226998)(16,0.221341)(17,0.224010)(18,0.217737)(19,0.224196)(20,0.221112)(21,0.223186)(22,0.219914)(23,0.219834)(24,0.215628)(25,0.219417)
                };
                \addplot[color=green!50!black, mark=triangle*, solid] coordinates {
                  (1,0.184236)(2,0.227552)(3,0.380152)(4,0.283251)(5,0.345750)(6,0.261890)(7,0.262195)(8,0.211822)(9,0.217818)(10,0.196917)(11,0.213187)(12,0.194701)(13,0.201835)(14,0.189879)(15,0.196563)(16,0.186753)(17,0.195606)(18,0.188213)(19,0.189710)(20,0.182406)(21,0.185527)(22,0.183549)(23,0.185145)(24,0.183151)(25,0.182375)
                };
                \addplot[color=red, mark=diamond*, solid] coordinates {
                  (1,0.125024)(2,0.170202)(3,0.194745)(4,0.215684)(5,0.196738)(6,0.212470)(7,0.194990)(8,0.210475)(9,0.200847)(10,0.210021)(11,0.204435)(12,0.209302)(13,0.202878)(14,0.204267)(15,0.202655)(16,0.201196)(17,0.201114)(18,0.203619)(19,0.199165)(20,0.197882)(21,0.206127)(22,0.198705)(23,0.198091)(24,0.200712)(25,0.200771)
                };
                \addplot[color=blue, mark=diamond*, solid] coordinates {
                  (1,0.136549)(2,0.192043)(3,0.215561)(4,0.211532)(5,0.197305)(6,0.211792)(7,0.194791)(8,0.206724)(9,0.199770)(10,0.196517)(11,0.198687)(12,0.199492)(13,0.197134)(14,0.193457)(15,0.197038)(16,0.192137)(17,0.200647)(18,0.196123)(19,0.198094)(20,0.188828)(21,0.197757)(22,0.188626)(23,0.194023)(24,0.191080)(25,0.194897)
                };
                \addplot[color=green!50!black, mark=diamond*, solid] coordinates {
                  (1,0.152152)(2,0.195569)(3,0.198118)(4,0.201586)(5,0.183329)(6,0.185172)(7,0.165884)(8,0.179778)(9,0.166307)(10,0.173282)(11,0.162447)(12,0.172290)(13,0.165435)(14,0.168055)(15,0.164492)(16,0.165533)(17,0.163234)(18,0.165340)(19,0.160925)(20,0.158658)(21,0.162066)(22,0.163691)(23,0.160300)(24,0.160014)(25,0.160786)
                };
            \end{axis}
        \end{tikzpicture}
    \end{minipage}}

    \vspace{0.4em}
    
    \subfloat{
        \begin{minipage}[t]{0.49\textwidth}
        \centering
        \begin{tikzpicture}[trim axis left, trim axis right]
            \begin{axis}[
                 width=\textwidth,
                 height=0.85\textwidth,
                 xlabel={Epoch},
                 xlabel style={yshift=4pt},
                 xtick={0,5,10,15,20,25},
                 xmin=-1.5, xmax=26.5,
                 ylabel={MS},
                 ytick={0.52, 0.53, 0.54, 0.55, 0.56, 0.57, 0.58, 0.59},
                 grid=both,
                 grid style={line width=.1pt, dashed, draw=gray!10},
                 major grid style={line width=.2pt,draw=gray!50},
                 label style={font=\scriptsize},
                 tick label style={font=\scriptsize},
                 every axis title/.style={at={(0.5,1.0)}, above, font=\small},
                 title={CLIP: TopKSAE},
                 mark size=0.8pt,
                 line width=0.4pt,
                 scaled y ticks=false,
                 yticklabel style={/pgf/number format/fixed, /pgf/number format/precision=4},
                 axis x line*=bottom,
                 axis y line*=left,
                 ]
                \addplot[color=red, mark=square*, solid] coordinates {
                  (1,0.518810)(2,0.555792)(3,0.558492)(4,0.547672)(5,0.551403)(6,0.535164)(7,0.543523)(8,0.551608)(9,0.556436)(10,0.543117)(11,0.567647)(12,0.552454)(13,0.558890)(14,0.566940)(15,0.564101)(16,0.563636)(17,0.567944)(18,0.576990)(19,0.575055)(20,0.569835)(21,0.575062)(22,0.564033)(23,0.571402)(24,0.567768)(25,0.570926)
                };
                \addplot[color=blue, mark=square*, solid] coordinates {
                  (1,0.559343)(2,0.562192)(3,0.560544)(4,0.565466)(5,0.553417)(6,0.564141)(7,0.570791)(8,0.546334)(9,0.558396)(10,0.548036)(11,0.547470)(12,0.560211)(13,0.558804)(14,0.542578)(15,0.549195)(16,0.547107)(17,0.560068)(18,0.558926)(19,0.567978)(20,0.564727)(21,0.571904)(22,0.566190)(23,0.572089)(24,0.576162)(25,0.563852)
                };
                \addplot[color=green!50!black, mark=square*, solid] coordinates {
                  (1,0.590923)(2,0.568869)(3,0.569229)(4,0.570096)(5,0.580476)(6,0.571285)(7,0.575390)(8,0.567881)(9,0.578989)(10,0.562699)(11,0.575796)(12,0.577039)(13,0.569963)(14,0.577828)(15,0.575206)(16,0.577289)(17,0.569712)(18,0.563349)(19,0.578962)(20,0.574584)(21,0.567285)(22,0.576437)(23,0.573115)(24,0.573260)(25,0.572445)
                };
                \addplot[color=red, mark=triangle*, solid] coordinates {
                  (1,0.519637)(2,0.542371)(3,0.544607)(4,0.535862)(5,0.543129)(6,0.546035)(7,0.549890)(8,0.556565)(9,0.567566)(10,0.567405)(11,0.561616)(12,0.564776)(13,0.568986)(14,0.570702)(15,0.571013)(16,0.564854)(17,0.568601)(18,0.573914)(19,0.575929)(20,0.570054)(21,0.575523)(22,0.575532)(23,0.577276)(24,0.577846)(25,0.566893)
                };
                \addplot[color=blue, mark=triangle*, solid] coordinates {
                  (1,0.526980)(2,0.540647)(3,0.551105)(4,0.554930)(5,0.560891)(6,0.559586)(7,0.558750)(8,0.554334)(9,0.560385)(10,0.559636)(11,0.562078)(12,0.555027)(13,0.559096)(14,0.561466)(15,0.568801)(16,0.561863)(17,0.568469)(18,0.568165)(19,0.569131)(20,0.568961)(21,0.575872)(22,0.569862)(23,0.573349)(24,0.568969)(25,0.574707)
                };
                \addplot[color=green!50!black, mark=triangle*, solid] coordinates {
                  (1,0.554833)(2,0.557688)(3,0.558038)(4,0.553763)(5,0.555909)(6,0.554941)(7,0.550438)(8,0.546278)(9,0.551650)(10,0.541117)(11,0.557090)(12,0.555372)(13,0.552432)(14,0.559365)(15,0.556913)(16,0.561071)(17,0.566441)(18,0.565018)(19,0.566353)(20,0.570115)(21,0.568168)(22,0.567654)(23,0.563833)(24,0.571124)(25,0.568264)
                };
                \addplot[color=red, mark=diamond*, solid] coordinates {
                  (1,0.517115)(2,0.534710)(3,0.544541)(4,0.551169)(5,0.551476)(6,0.558022)(7,0.558683)(8,0.559498)(9,0.558689)(10,0.559295)(11,0.561177)(12,0.559599)(13,0.561824)(14,0.562897)(15,0.561579)(16,0.557735)(17,0.561593)(18,0.563210)(19,0.561387)(20,0.561993)(21,0.560715)(22,0.561452)(23,0.564333)(24,0.561531)(25,0.564859)
                };
                \addplot[color=blue, mark=diamond*, solid] coordinates {
                  (1,0.529364)(2,0.532262)(3,0.542103)(4,0.543602)(5,0.552204)(6,0.554918)(7,0.554888)(8,0.557176)(9,0.560217)(10,0.555443)(11,0.558185)(12,0.558048)(13,0.558299)(14,0.558223)(15,0.561621)(16,0.560725)(17,0.563470)(18,0.562216)(19,0.566957)(20,0.560981)(21,0.567810)(22,0.561431)(23,0.565516)(24,0.560065)(25,0.561639)
                };
                \addplot[color=green!50!black, mark=diamond*, solid] coordinates {
                  (1,0.530684)(2,0.536842)(3,0.544882)(4,0.543730)(5,0.546452)(6,0.543891)(7,0.546469)(8,0.551253)(9,0.543321)(10,0.549606)(11,0.548315)(12,0.556333)(13,0.550309)(14,0.551757)(15,0.548290)(16,0.554366)(17,0.549987)(18,0.551328)(19,0.552620)(20,0.555703)(21,0.557668)(22,0.559096)(23,0.555706)(24,0.554928)(25,0.557346)
                };
            \end{axis}
        \end{tikzpicture}
    \end{minipage}}\hspace{-4em}%
    \subfloat{
        \begin{minipage}[t]{0.49\textwidth}
        \centering
        \begin{tikzpicture}[trim axis left, trim axis right]
            \begin{axis}[
                 width=\textwidth,
                 height=0.85\textwidth,
                 xlabel={Epoch},
                 xlabel style={yshift=4pt},
                 xtick={0,5,10,15,20,25},
                 xmin=-1.5, xmax=26.5,
                 ylabel={},
                 ylabel style={yshift=-3pt},
                 grid=both,
                 grid style={line width=.1pt, dashed, draw=gray!10},
                 major grid style={line width=.2pt,draw=gray!50},
                 label style={font=\scriptsize},
                 tick label style={font=\scriptsize},
                 every axis title/.style={at={(0.5,1.0)}, above, font=\small},
                 title={CLIP: BatchTopKSAE},
                 mark size=0.8pt,
                 line width=0.4pt,
                 scaled y ticks=false,
                 yticklabel style={/pgf/number format/fixed zerofill, /pgf/number format/precision=2},
                 axis x line*=bottom,
                 axis y line*=right,
                 ]
                \addplot[color=red, mark=square*, solid] coordinates {
                  (1,0.641378)(2,0.680633)(3,0.644084)(4,0.624769)(5,0.619735)(6,0.629092)(7,0.626608)(8,0.627240)(9,0.625200)(10,0.612213)(11,0.610507)(12,0.611730)(13,0.593398)(14,0.578336)(15,0.575793)(16,0.569459)(17,0.574763)(18,0.573856)(19,0.571715)(20,0.571026)(21,0.574715)(22,0.572173)(23,0.567991)(24,0.571485)(25,0.571182)
                };
                \addplot[color=blue, mark=square*, solid] coordinates {
                  (1,0.660630)(2,0.691341)(3,0.701017)(4,0.698275)(5,0.679961)(6,0.657375)(7,0.650669)(8,0.641362)(9,0.641750)(10,0.623836)(11,0.614439)(12,0.594953)(13,0.593624)(14,0.584102)(15,0.583352)(16,0.576152)(17,0.576240)(18,0.580950)(19,0.567162)(20,0.570437)(21,0.564201)(22,0.562848)(23,0.564705)(24,0.563953)(25,0.566480)
                };
                \addplot[color=green!50!black, mark=square*, solid] coordinates {
                  (1,0.672124)(2,0.654097)(3,0.657674)(4,0.620112)(5,0.646611)(6,0.661910)(7,0.668821)(8,0.668855)(9,0.660604)(10,0.643178)(11,0.644187)(12,0.631472)(13,0.623813)(14,0.626637)(15,0.611460)(16,0.609304)(17,0.598112)(18,0.591580)(19,0.583368)(20,0.573086)(21,0.574499)(22,0.578599)(23,0.576539)(24,0.576688)(25,0.574314)
                };
                \addplot[color=red, mark=triangle*, solid] coordinates {
                  (1,0.590617)(2,0.590662)(3,0.592891)(4,0.570424)(5,0.564460)(6,0.564109)(7,0.559988)(8,0.559373)(9,0.557236)(10,0.557387)(11,0.560844)(12,0.563465)(13,0.570633)(14,0.572093)(15,0.573088)(16,0.563942)(17,0.564639)(18,0.563663)(19,0.567871)(20,0.567498)(21,0.567379)(22,0.569060)(23,0.570862)(24,0.566012)(25,0.564163)
                };
                \addplot[color=blue, mark=triangle*, solid] coordinates {
                  (1,0.596996)(2,0.592898)(3,0.600834)(4,0.589465)(5,0.573102)(6,0.565056)(7,0.559506)(8,0.559455)(9,0.563668)(10,0.561962)(11,0.564232)(12,0.565183)(13,0.559941)(14,0.569078)(15,0.567761)(16,0.563601)(17,0.563648)(18,0.558787)(19,0.555830)(20,0.551411)(21,0.558508)(22,0.555312)(23,0.555779)(24,0.556067)(25,0.557259)
                };
                \addplot[color=green!50!black, mark=triangle*, solid] coordinates {
                  (1,0.617318)(2,0.604637)(3,0.596737)(4,0.584895)(5,0.569796)(6,0.560070)(7,0.552412)(8,0.554892)(9,0.557975)(10,0.556718)(11,0.561742)(12,0.563075)(13,0.558568)(14,0.565104)(15,0.559755)(16,0.563637)(17,0.559760)(18,0.561012)(19,0.556322)(20,0.556299)(21,0.554636)(22,0.552434)(23,0.549638)(24,0.552398)(25,0.552294)
                };
                \addplot[color=red, mark=diamond*, solid] coordinates {
                  (1,0.544122)(2,0.544145)(3,0.548506)(4,0.547869)(5,0.545219)(6,0.547331)(7,0.546451)(8,0.546600)(9,0.545907)(10,0.546057)(11,0.546805)(12,0.548668)(13,0.545141)(14,0.551564)(15,0.549685)(16,0.550779)(17,0.552621)(18,0.548395)(19,0.545973)(20,0.552173)(21,0.547656)(22,0.549996)(23,0.549587)(24,0.549498)(25,0.544986)
                };
                \addplot[color=blue, mark=diamond*, solid] coordinates {
                  (1,0.543110)(2,0.543246)(3,0.546570)(4,0.545182)(5,0.544326)(6,0.545769)(7,0.542325)(8,0.544558)(9,0.542486)(10,0.545237)(11,0.540971)(12,0.546995)(13,0.543610)(14,0.547566)(15,0.547602)(16,0.549172)(17,0.547659)(18,0.545215)(19,0.545532)(20,0.543965)(21,0.545033)(22,0.547020)(23,0.543740)(24,0.544813)(25,0.546712)
                };
                \addplot[color=green!50!black, mark=diamond*, solid] coordinates {
                  (1,0.542362)(2,0.536719)(3,0.540486)(4,0.538350)(5,0.535505)(6,0.535326)(7,0.538336)(8,0.534971)(9,0.534775)(10,0.536916)(11,0.536822)(12,0.537121)(13,0.532882)(14,0.535964)(15,0.534578)(16,0.536493)(17,0.535644)(18,0.533127)(19,0.535961)(20,0.535249)(21,0.536014)(22,0.534109)(23,0.539625)(24,0.537397)(25,0.538075)
                };
            \end{axis}
        \end{tikzpicture}
    \end{minipage}}

    \vspace{0.4em}

    \subfloat{
        \begin{minipage}[t]{0.49\textwidth}
        \centering
        \begin{tikzpicture}[trim axis left, trim axis right]
            \begin{axis}[
                 width=\textwidth,
                 height=0.85\textwidth,
                 xlabel={Epoch},
                 xlabel style={yshift=4pt},
                 xtick={0,5,10,15,20,25},
                 xmin=-1.5, xmax=26.5,
                 ylabel={MS},
                 grid=both,
                 grid style={line width=.1pt, dashed, draw=gray!10},
                 major grid style={line width=.2pt,draw=gray!50},
                 label style={font=\scriptsize},
                 tick label style={font=\scriptsize},
                 every axis title/.style={at={(0.5,1.0)}, above, font=\small},
                 title={BLIP: TopKSAE},
                 mark size=0.8pt,
                 line width=0.4pt,
                 scaled y ticks=false,
                 yticklabel style={/pgf/number format/fixed zerofill, /pgf/number format/precision=2},
                 axis x line*=bottom,
                 axis y line*=left,
                 ]
                \addplot[color=red, mark=square*, solid] coordinates {
                  (1,0.443657)(2,0.484634)(3,0.495193)(4,0.475261)(5,0.471149)(6,0.474731)(7,0.453457)(8,0.458494)(9,0.435431)(10,0.438883)(11,0.437457)(12,0.433803)(13,0.439068)(14,0.431164)(15,0.431202)(16,0.426548)(17,0.427048)(18,0.421251)(19,0.425101)(20,0.427937)(21,0.425974)(22,0.433753)(23,0.432047)(24,0.423934)(25,0.433559)
                };
                \addplot[color=blue, mark=square*, solid] coordinates {
                  (1,0.460681)(2,0.478833)(3,0.496211)(4,0.492703)(5,0.504080)(6,0.489734)(7,0.486868)(8,0.469550)(9,0.464658)(10,0.456144)(11,0.465484)(12,0.465749)(13,0.459940)(14,0.477809)(15,0.458114)(16,0.448556)(17,0.460628)(18,0.462951)(19,0.455720)(20,0.457828)(21,0.446463)(22,0.452054)(23,0.454444)(24,0.456964)(25,0.440146)
                };
                \addplot[color=green!50!black, mark=square*, solid] coordinates {
                  (1,0.490104)(2,0.514378)(3,0.556569)(4,0.541557)(5,0.535570)(6,0.525500)(7,0.531243)(8,0.518233)(9,0.518752)(10,0.514039)(11,0.515165)(12,0.506489)(13,0.492824)(14,0.498273)(15,0.491630)(16,0.490878)(17,0.491368)(18,0.475377)(19,0.480102)(20,0.477137)(21,0.469954)(22,0.479634)(23,0.466708)(24,0.476746)(25,0.473375)
                };
                \addplot[color=red, mark=triangle*, solid] coordinates {
                  (1,0.441092)(2,0.474236)(3,0.443548)(4,0.446129)(5,0.437531)(6,0.431239)(7,0.435617)(8,0.430506)(9,0.427833)(10,0.427733)(11,0.431419)(12,0.426402)(13,0.431702)(14,0.433661)(15,0.425539)(16,0.430659)(17,0.429492)(18,0.428478)(19,0.431371)(20,0.423839)(21,0.435669)(22,0.429057)(23,0.426730)(24,0.428172)(25,0.429164)
                };
                \addplot[color=blue, mark=triangle*, solid] coordinates {
                  (1,0.471625)(2,0.503394)(3,0.457337)(4,0.476211)(5,0.465015)(6,0.462741)(7,0.439636)(8,0.454758)(9,0.457097)(10,0.442207)(11,0.444378)(12,0.431249)(13,0.445495)(14,0.434104)(15,0.435980)(16,0.438815)(17,0.447658)(18,0.435607)(19,0.444104)(20,0.438780)(21,0.443336)(22,0.441168)(23,0.439346)(24,0.438611)(25,0.440591)
                };
                \addplot[color=green!50!black, mark=triangle*, solid] coordinates {
                  (1,0.501974)(2,0.525872)(3,0.484348)(4,0.498956)(5,0.483423)(6,0.487257)(7,0.471584)(8,0.492637)(9,0.468000)(10,0.486547)(11,0.469311)(12,0.465663)(13,0.460358)(14,0.475903)(15,0.461294)(16,0.460062)(17,0.460257)(18,0.454997)(19,0.462315)(20,0.455416)(21,0.451434)(22,0.457064)(23,0.455609)(24,0.450646)(25,0.464056)
                };
                \addplot[color=red, mark=diamond*, solid] coordinates {
                  (1,0.436500)(2,0.452735)(3,0.426446)(4,0.454376)(5,0.424942)(6,0.433779)(7,0.434966)(8,0.437665)(9,0.436205)(10,0.437362)(11,0.426282)(12,0.435327)(13,0.425429)(14,0.428653)(15,0.411911)(16,0.423012)(17,0.409700)(18,0.413840)(19,0.418860)(20,0.420522)(21,0.412060)(22,0.422476)(23,0.411400)(24,0.415720)(25,0.415706)
                };
                \addplot[color=blue, mark=diamond*, solid] coordinates {
                  (1,0.469075)(2,0.491255)(3,0.425730)(4,0.467933)(5,0.429789)(6,0.447972)(7,0.444765)(8,0.455910)(9,0.432850)(10,0.444168)(11,0.433956)(12,0.444305)(13,0.433919)(14,0.440222)(15,0.437113)(16,0.429108)(17,0.425583)(18,0.431032)(19,0.429123)(20,0.428771)(21,0.415553)(22,0.421545)(23,0.423027)(24,0.417550)(25,0.411493)
                };
                \addplot[color=green!50!black, mark=diamond*, solid] coordinates {
                  (1,0.498613)(2,0.520408)(3,0.455248)(4,0.490761)(5,0.449017)(6,0.459365)(7,0.453512)(8,0.462120)(9,0.453775)(10,0.455805)(11,0.453526)(12,0.470711)(13,0.451714)(14,0.455936)(15,0.446339)(16,0.460834)(17,0.440513)(18,0.459079)(19,0.436194)(20,0.444562)(21,0.437838)(22,0.448377)(23,0.427228)(24,0.439861)(25,0.436096)
                };
            \end{axis}
        \end{tikzpicture}
    \end{minipage}}\hspace{-4em}%
    \subfloat{
        \begin{minipage}[t]{0.49\textwidth}
        \centering
        \begin{tikzpicture}[trim axis left, trim axis right]
            \begin{axis}[
                 width=\textwidth,
                 height=0.85\textwidth,
                 xlabel={Epoch},
                 xlabel style={yshift=4pt},
                 xtick={0,5,10,15,20,25},
                 xmin=-1.5, xmax=26.5,
                 ylabel={},
                 ytick={0.35, 0.40, 0.45, 0.50, 0.55, 0.60, 0.65, 0.7},
                 grid=both,
                 grid style={line width=.1pt, dashed, draw=gray!10},
                 major grid style={line width=.2pt,draw=gray!50},
                 label style={font=\scriptsize},
                 tick label style={font=\scriptsize},
                 every axis title/.style={at={(0.5,1.0)}, above, font=\small},
                 title={BLIP: BatchTopKSAE},
                 mark size=0.8pt,
                 line width=0.4pt,
                 scaled y ticks=false,
                 yticklabel style={/pgf/number format/fixed zerofill, /pgf/number format/precision=2},
                 axis y line*=right,
                 axis x line*=bottom,
                 ]
                \addplot[color=red, mark=square*, solid] coordinates {
                  (1,0.606044)(2,0.592937)(3,0.565957)(4,0.552771)(5,0.535170)(6,0.508782)(7,0.470582)(8,0.443529)(9,0.428349)(10,0.423027)(11,0.432272)(12,0.426835)(13,0.433247)(14,0.421220)(15,0.422179)(16,0.425871)(17,0.420798)(18,0.417110)(19,0.420064)(20,0.423742)(21,0.425674)(22,0.428066)(23,0.417623)(24,0.426189)(25,0.422952)
                };
                \addplot[color=blue, mark=square*, solid] coordinates {
                  (1,0.624054)(2,0.611228)(3,0.599248)(4,0.592781)(5,0.567208)(6,0.549346)(7,0.548554)(8,0.531947)(9,0.526297)(10,0.511055)(11,0.504604)(12,0.488479)(13,0.485274)(14,0.472376)(15,0.469844)(16,0.458703)(17,0.457558)(18,0.452991)(19,0.448920)(20,0.466324)(21,0.456357)(22,0.446051)(23,0.438010)(24,0.431219)(25,0.425188)
                };
                \addplot[color=green!50!black, mark=square*, solid] coordinates {
                  (1,0.642942)(2,0.629927)(3,0.601422)(4,0.595149)(5,0.574933)(6,0.569777)(7,0.557349)(8,0.543572)(9,0.536468)(10,0.529505)(11,0.512946)(12,0.512630)(13,0.506516)(14,0.488574)(15,0.480110)(16,0.465628)(17,0.464105)(18,0.461013)(19,0.476011)(20,0.469862)(21,0.479786)(22,0.476653)(23,0.489318)(24,0.489932)(25,0.503171)
                };
                \addplot[color=red, mark=triangle*, solid] coordinates {
                  (1,0.535136)(2,0.512252)(3,0.467238)(4,0.428191)(5,0.413324)(6,0.410923)(7,0.402083)(8,0.393165)(9,0.398245)(10,0.396069)(11,0.395514)(12,0.390466)(13,0.400008)(14,0.392396)(15,0.384948)(16,0.387902)(17,0.385641)(18,0.376909)(19,0.378827)(20,0.374982)(21,0.377368)(22,0.372051)(23,0.379263)(24,0.369278)(25,0.374396)
                };
                \addplot[color=blue, mark=triangle*, solid] coordinates {
                  (1,0.540338)(2,0.524318)(3,0.488004)(4,0.446964)(5,0.420454)(6,0.410525)(7,0.397763)(8,0.392009)(9,0.388326)(10,0.383202)(11,0.379623)(12,0.378391)(13,0.379749)(14,0.376463)(15,0.378213)(16,0.376006)(17,0.370304)(18,0.366824)(19,0.368249)(20,0.359198)(21,0.361951)(22,0.357426)(23,0.361824)(24,0.356980)(25,0.362066)
                };
                \addplot[color=green!50!black, mark=triangle*, solid] coordinates {
                  (1,0.563685)(2,0.550867)(3,0.522604)(4,0.475830)(5,0.447715)(6,0.404227)(7,0.395926)(8,0.394806)(9,0.394569)(10,0.392791)(11,0.399392)(12,0.399182)(13,0.393927)(14,0.393103)(15,0.385885)(16,0.373617)(17,0.370144)(18,0.365011)(19,0.367825)(20,0.366689)(21,0.365559)(22,0.360170)(23,0.363799)(24,0.357945)(25,0.358550)
                };
                \addplot[color=red, mark=diamond*, solid] coordinates {
                  (1,0.457108)(2,0.447829)(3,0.429409)(4,0.410518)(5,0.407874)(6,0.400454)(7,0.402618)(8,0.402586)(9,0.396041)(10,0.395880)(11,0.396376)(12,0.390862)(13,0.388167)(14,0.390138)(15,0.386911)(16,0.382851)(17,0.379613)(18,0.376532)(19,0.375599)(20,0.382191)(21,0.384785)(22,0.377126)(23,0.376207)(24,0.375471)(25,0.376702)
                };
                \addplot[color=blue, mark=diamond*, solid] coordinates {
                  (1,0.472345)(2,0.453504)(3,0.436089)(4,0.405687)(5,0.401930)(6,0.388792)(7,0.395179)(8,0.384150)(9,0.385849)(10,0.383126)(11,0.376042)(12,0.380780)(13,0.372432)(14,0.374642)(15,0.369848)(16,0.369540)(17,0.365493)(18,0.372344)(19,0.361641)(20,0.364721)(21,0.362111)(22,0.361340)(23,0.363670)(24,0.366718)(25,0.362287)
                };
                \addplot[color=green!50!black, mark=diamond*, solid] coordinates {
                  (1,0.504518)(2,0.477291)(3,0.450359)(4,0.410031)(5,0.407082)(6,0.391546)(7,0.389201)(8,0.380984)(9,0.381471)(10,0.378124)(11,0.374617)(12,0.373529)(13,0.366951)(14,0.369347)(15,0.363490)(16,0.368614)(17,0.364090)(18,0.364817)(19,0.360256)(20,0.361860)(21,0.361879)(22,0.361802)(23,0.357137)(24,0.364583)(25,0.360588)
                };
            \end{axis}
        \end{tikzpicture}
    \end{minipage}}
    \vspace{0.3em}
    \ref{legendTopK}\par\vspace{0.2em}
    \ref{legendBatch}

    \caption{Epoch evolution of MS for TopKSAE (left column) and BatchTopKSAE (right column) across DINO, CLIP, and BLIP models.}
    \label{fig:epoch_evolution_ms_score}
\end{figure*}
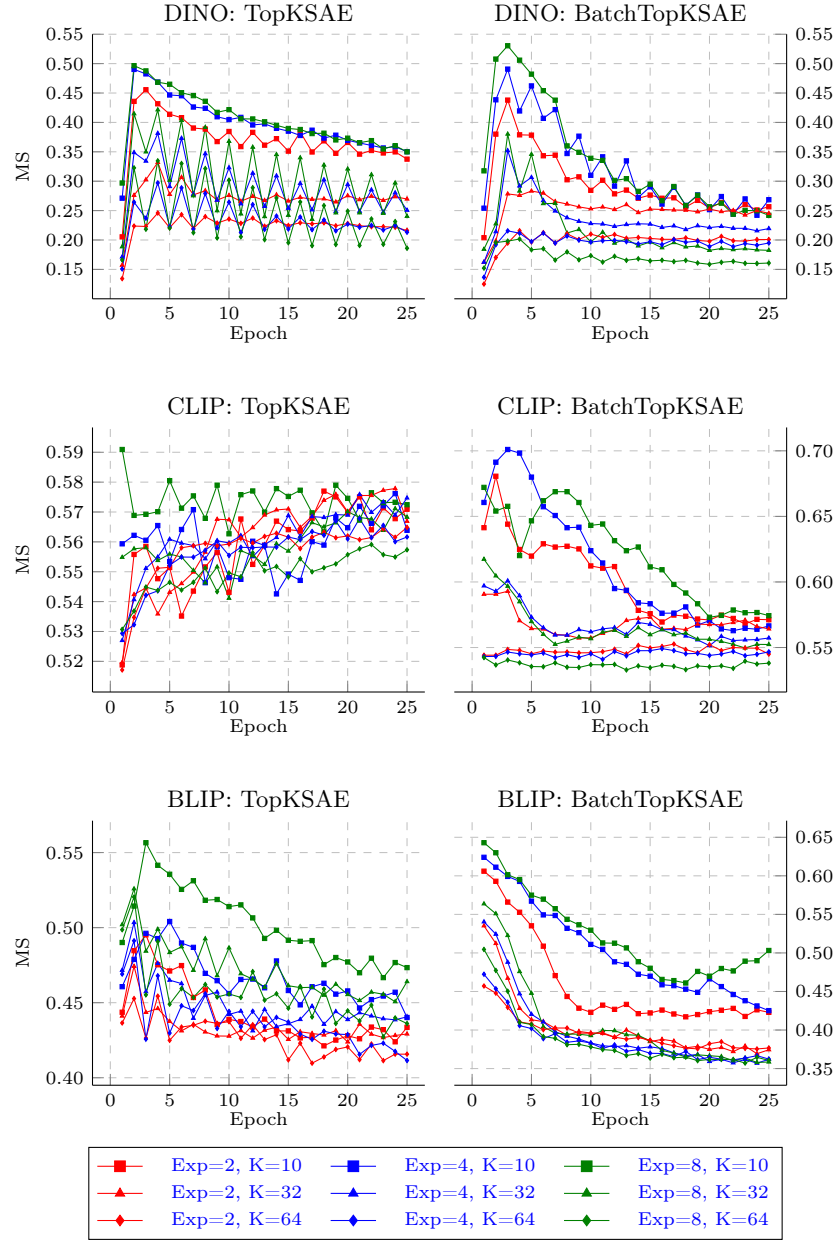

%% file: FIGS/scattered/epoch_tversky_similarity.tex
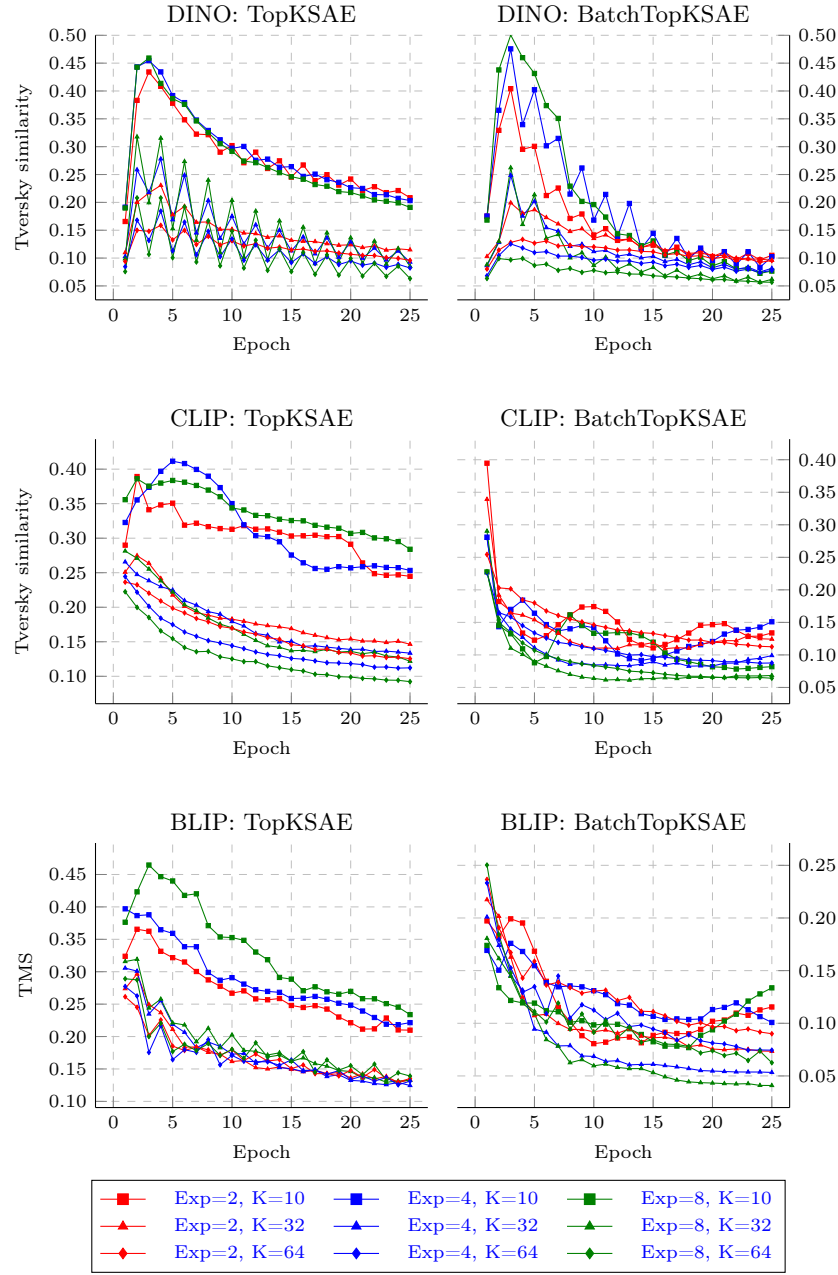
\begin{figure*}[h]
    \centering

    \subfloat{
        \begin{minipage}[t]{0.49\textwidth}
        \centering
        \begin{tikzpicture}[trim axis left, trim axis right]
            \begin{axis}[
                 width=\textwidth,
                 height=0.85\textwidth,
                 xlabel={Epoch},
                 xtick={0,5,10,15,20,25},
                 xmin=-1.5, xmax=26.5,
                 ylabel={Tversky similarity},
                 grid=both,
                 grid style={line width=.1pt, dashed, draw=gray!10},
                 major grid style={line width=.2pt,draw=gray!50},
                 label style={font=\scriptsize},
                 tick label style={font=\scriptsize},
                 every axis title/.style={at={(0.5,1.0)}, above, font=\small},
                 title={DINO: TopKSAE},
                 mark size=1.2pt,
                 line width=0.6pt,
                 scaled y ticks=false,
                 yticklabel style={
                     /pgf/number format/fixed,
                     /pgf/number format/fixed zerofill,
                     /pgf/number format/precision=2
                 },
                 legend to name=legendTopK,
                 legend style={legend columns=3, font=\scriptsize, column sep=0.3cm},
                 axis y line*=left,
                 axis x line*=bottom,
                 mark size=0.8pt,
                 line width=0.4pt,
                 ytick={0.05, 0.1, 0.15,0.2,0.25,0.3,0.35,0.4,0.45,0.5},
                 ymin=0.025,
                 ymax=0.50
                 ]
                \addplot[color=red, mark=square*, solid] coordinates {
                  (1,0.165661)(2,0.383161)(3,0.434087)(4,0.408340)(5,0.378072)(6,0.348232)(7,0.322574)(8,0.321543)(9,0.290206)(10,0.301940)(11,0.271203)(12,0.290192)(13,0.260858)(14,0.274535)(15,0.245386)(16,0.267102)(17,0.238865)(18,0.249675)(19,0.231068)(20,0.241882)(21,0.221381)(22,0.228060)(23,0.217508)(24,0.221201)(25,0.208304)
                };
                \addlegendentry{Exp=2, K=10}
                \addplot[color=blue, mark=square*, solid] coordinates {
                  (1,0.191391)(2,0.443232)(3,0.454559)(4,0.434261)(5,0.391724)(6,0.379112)(7,0.347827)(8,0.329084)(9,0.312376)(10,0.297149)(11,0.300438)(12,0.275519)(13,0.277674)(14,0.263045)(15,0.264173)(16,0.246601)(17,0.250736)(18,0.240691)(19,0.235968)(20,0.226324)(21,0.224959)(22,0.214009)(23,0.213627)(24,0.207201)(25,0.203248)
                };
                \addlegendentry{Exp=4, K=10}
                \addplot[color=green!50!black, mark=square*, solid] coordinates {
                  (1,0.189893)(2,0.442769)(3,0.459115)(4,0.413603)(5,0.386292)(6,0.375990)(7,0.346244)(8,0.325302)(9,0.305276)(10,0.291782)(11,0.274218)(12,0.271308)(13,0.262598)(14,0.253427)(15,0.246162)(16,0.241605)(17,0.232043)(18,0.229080)(19,0.219420)(20,0.217726)(21,0.211617)(22,0.204455)(23,0.202108)(24,0.199118)(25,0.191029)
                };
                \addlegendentry{Exp=8, K=10}
                \addplot[color=red, mark=triangle*, solid] coordinates {
                  (1,0.109762)(2,0.200093)(3,0.216974)(4,0.230527)(5,0.177821)(6,0.192801)(7,0.164319)(8,0.164869)(9,0.151631)(10,0.151659)(11,0.144969)(12,0.143854)(13,0.137345)(14,0.139837)(15,0.132039)(16,0.130182)(17,0.129646)(18,0.125793)(19,0.122964)(20,0.123858)(21,0.118941)(22,0.121712)(23,0.114358)(24,0.116575)(25,0.114855)
                };
                \addlegendentry{Exp=2, K=32}
                \addplot[color=blue, mark=triangle*, solid] coordinates {
                  (1,0.102750)(2,0.258580)(3,0.217836)(4,0.278051)(5,0.168874)(6,0.248850)(7,0.145097)(8,0.203473)(9,0.135315)(10,0.175075)(11,0.125417)(12,0.159735)(13,0.118027)(14,0.149950)(15,0.112907)(16,0.137954)(17,0.107534)(18,0.136178)(19,0.101472)(20,0.128849)(21,0.099802)(22,0.117912)(23,0.096555)(24,0.112551)(25,0.092800)
                };
                \addlegendentry{Exp=4, K=32}
                \addplot[color=green!50!black, mark=triangle*, solid] coordinates {
                  (1,0.098626)(2,0.317589)(3,0.199369)(4,0.315317)(5,0.152636)(6,0.273399)(7,0.132020)(8,0.239938)(9,0.109734)(10,0.203507)(11,0.101554)(12,0.184397)(13,0.096886)(14,0.166823)(15,0.095040)(16,0.155519)(17,0.089877)(18,0.145334)(19,0.093761)(20,0.136442)(21,0.090541)(22,0.130399)(23,0.088986)(24,0.117026)(25,0.085063)
                };
                \addlegendentry{Exp=8, K=32}
                \addplot[color=red, mark=diamond*, solid] coordinates {
                  (1,0.094000)(2,0.149885)(3,0.148073)(4,0.158731)(5,0.132547)(6,0.149807)(7,0.124440)(8,0.139139)(9,0.123597)(10,0.130929)(11,0.121772)(12,0.125758)(13,0.117645)(14,0.119748)(15,0.114715)(16,0.116125)(17,0.112194)(18,0.112819)(19,0.109187)(20,0.106870)(21,0.104759)(22,0.104565)(23,0.100957)(24,0.099460)(25,0.095918)
                };
                \addlegendentry{Exp=2, K=64}
                \addplot[color=blue, mark=diamond*, solid] coordinates {
                  (1,0.084578)(2,0.168032)(3,0.131703)(4,0.184297)(5,0.112931)(6,0.164556)(7,0.105795)(8,0.148178)(9,0.102577)(10,0.134516)(11,0.096089)(12,0.123752)(13,0.096271)(14,0.115058)(15,0.091996)(16,0.107393)(17,0.090333)(18,0.102252)(19,0.088384)(20,0.094831)(21,0.087237)(22,0.090979)(23,0.083444)(24,0.088698)(25,0.082247)
                };
                \addlegendentry{Exp=4, K=64}
                \addplot[color=green!50!black, mark=diamond*, solid] coordinates {
                  (1,0.075761)(2,0.208044)(3,0.106680)(4,0.208051)(5,0.100589)(6,0.192071)(7,0.092947)(8,0.166824)(9,0.086023)(10,0.148262)(11,0.082048)(12,0.132122)(13,0.077916)(14,0.116353)(15,0.075655)(16,0.110241)(17,0.070905)(18,0.105693)(19,0.069497)(20,0.099340)(21,0.067740)(22,0.090539)(23,0.067196)(24,0.085527)(25,0.063281)
                };
                \addlegendentry{Exp=8, K=64}
            \end{axis}
        \end{tikzpicture}
   \end{minipage}}\hspace{-4em}%
    \subfloat{
        \begin{minipage}[t]{0.49\textwidth}
        \centering
        \begin{tikzpicture}[trim axis left, trim axis right]
            \begin{axis}[
                 width=\textwidth,
                 height=0.85\textwidth,
                 xlabel={Epoch},
                 xtick={0,5,10,15,20,25},
                 xmin=-1.5, xmax=26.5,
                 ylabel={},
                 grid=both,
                 grid style={line width=.1pt, dashed, draw=gray!10},
                 major grid style={line width=.2pt,draw=gray!50},
                 label style={font=\scriptsize},
                 tick label style={font=\scriptsize},
                 every axis title/.style={at={(0.5,1.0)}, above, font=\small},
                 title={DINO: BatchTopKSAE},
                 mark size=1.2pt,
                 line width=0.6pt,
                 scaled y ticks=false,
                 yticklabel style={
                     /pgf/number format/fixed,
                     /pgf/number format/fixed zerofill,
                     /pgf/number format/precision=2
                 },
                 legend to name=legendBatch,
                 legend style={legend columns=3, font=\scriptsize, column sep=0.3cm},
                 axis y line*=right,
                 axis x line*=bottom,
                 mark size=0.8pt,
                 line width=0.4pt,
                 ytick={0.05, 0.1, 0.15,0.2,0.25,0.3,0.35,0.4,0.45,0.5},
                 ymin=0.025,
                 ymax=0.50
                 ]
                \addplot[color=red, mark=square*, solid] coordinates {
                  (1,0.175275)(2,0.329336)(3,0.404179)(4,0.295371)(5,0.300644)(6,0.212280)(7,0.225692)(8,0.171125)(9,0.179094)(10,0.141447)(11,0.152761)(12,0.132503)(13,0.134137)(14,0.119277)(15,0.123228)(16,0.111553)(17,0.119578)(18,0.103470)(19,0.114272)(20,0.102773)(21,0.111038)(22,0.097280)(23,0.108649)(24,0.096844)(25,0.104949)
                };
                \addplot[color=blue, mark=square*, solid] coordinates {
                  (1,0.175506)(2,0.365382)(3,0.475636)(4,0.339609)(5,0.402205)(6,0.301553)(7,0.314846)(8,0.214951)(9,0.261705)(10,0.168166)(11,0.214265)(12,0.138947)(13,0.198058)(14,0.117263)(15,0.144423)(16,0.104797)(17,0.135240)(18,0.098872)(19,0.117905)(20,0.095535)(21,0.111561)(22,0.088150)(23,0.111431)(24,0.083310)(25,0.102520)
                };
                \addplot[color=green!50!black, mark=square*, solid] coordinates {
                  (1,0.168276)(2,0.437803)(3,0.500495)(4,0.459810)(5,0.431479)(6,0.373755)(7,0.350733)(8,0.229147)(9,0.201721)(10,0.196032)(11,0.173657)(12,0.143591)(13,0.140537)(14,0.122131)(15,0.130725)(16,0.105556)(17,0.111602)(18,0.095384)(19,0.101813)(20,0.086061)(21,0.093700)(22,0.080440)(23,0.081217)(24,0.072631)(25,0.076625)
                };
                \addplot[color=red, mark=triangle*, solid] coordinates {
                  (1,0.102684)(2,0.129707)(3,0.199160)(4,0.180400)(5,0.186779)(6,0.172753)(7,0.161349)(8,0.147732)(9,0.152601)(10,0.135128)(11,0.141551)(12,0.129851)(13,0.134362)(14,0.118730)(15,0.126622)(16,0.113336)(17,0.117512)(18,0.108729)(19,0.109274)(20,0.104420)(21,0.104777)(22,0.100016)(23,0.098456)(24,0.098099)(25,0.096057)
                };
                \addplot[color=blue, mark=triangle*, solid] coordinates {
                  (1,0.086477)(2,0.128081)(3,0.248498)(4,0.175417)(5,0.201858)(6,0.154590)(7,0.148383)(8,0.121201)(9,0.124790)(10,0.111614)(11,0.113689)(12,0.103616)(13,0.106849)(14,0.100404)(15,0.103105)(16,0.093768)(17,0.097568)(18,0.087999)(19,0.093444)(20,0.082578)(21,0.088842)(22,0.079804)(23,0.083767)(24,0.075025)(25,0.081641)
                };
                \addplot[color=green!50!black, mark=triangle*, solid] coordinates {
                  (1,0.088086)(2,0.129654)(3,0.262183)(4,0.160332)(5,0.213693)(6,0.135623)(7,0.142286)(8,0.100314)(9,0.109479)(10,0.086741)(11,0.101713)(12,0.079813)(13,0.089337)(14,0.074836)(15,0.083108)(16,0.067929)(17,0.078258)(18,0.065622)(19,0.071244)(20,0.062735)(21,0.068032)(22,0.058678)(23,0.065867)(24,0.056249)(25,0.061882)
                };
                \addplot[color=red, mark=diamond*, solid] coordinates {
                  (1,0.080316)(2,0.114320)(3,0.128111)(4,0.133557)(5,0.126806)(6,0.130577)(7,0.121766)(8,0.123019)(9,0.120352)(10,0.120122)(11,0.118336)(12,0.114295)(13,0.114609)(14,0.111830)(15,0.111384)(16,0.106066)(17,0.107942)(18,0.102749)(19,0.104502)(20,0.098949)(21,0.102049)(22,0.096230)(23,0.098900)(24,0.092350)(25,0.095587)
                };
                \addplot[color=blue, mark=diamond*, solid] coordinates {
                  (1,0.068204)(2,0.104585)(3,0.125380)(4,0.118189)(5,0.109833)(6,0.111443)(7,0.103507)(8,0.103573)(9,0.101098)(10,0.096097)(11,0.098123)(12,0.094580)(13,0.094956)(14,0.090397)(15,0.092828)(16,0.086156)(17,0.089075)(18,0.083218)(19,0.086729)(20,0.079643)(21,0.083770)(22,0.076380)(23,0.080143)(24,0.074213)(25,0.077746)
                };
                \addplot[color=green!50!black, mark=diamond*, solid] coordinates {
                  (1,0.063421)(2,0.099134)(3,0.096737)(4,0.099303)(5,0.086766)(6,0.089106)(7,0.077965)(8,0.081394)(9,0.074324)(10,0.078002)(11,0.073708)(12,0.074869)(13,0.071412)(14,0.071037)(15,0.068197)(16,0.067030)(17,0.065938)(18,0.064236)(19,0.062927)(20,0.060970)(21,0.061419)(22,0.059175)(23,0.058398)(24,0.056929)(25,0.056630)
                };
            \end{axis}
        \end{tikzpicture}
    \end{minipage}}\hfill%

    \vspace{0.4em}

    \subfloat{
        \begin{minipage}[t]{0.49\textwidth}
        \centering
        \begin{tikzpicture}[trim axis left, trim axis right]
            \begin{axis}[
                 width=\textwidth,
                 height=0.85\textwidth,
                 xlabel={Epoch},
                 xtick={0,5,10,15,20,25},
                 xmin=-1.5, xmax=26.5,
                 ylabel={Tversky similarity},
                 grid=both,
                 grid style={line width=.1pt, dashed, draw=gray!10},
                 major grid style={line width=.2pt,draw=gray!50},
                 label style={font=\scriptsize},
                 tick label style={font=\scriptsize},
                 every axis title/.style={at={(0.5,1.0)}, above, font=\small},
                 title={CLIP: TopKSAE},
                 mark size=1.2pt,
                 line width=0.6pt,
                 scaled y ticks=false,
                 yticklabel style={
                     /pgf/number format/fixed,
                     /pgf/number format/fixed zerofill,
                     /pgf/number format/precision=2
                 },
                 axis y line*=left,
                 axis x line*=bottom,
                 mark size=0.8pt,
                 line width=0.4pt,
                 ytick={0.05, 0.1, 0.15,0.2,0.25,0.3,0.35,0.4,0.45,0.5},
                 ]
                \addplot[color=red, mark=square*, solid] coordinates {
                  (1,0.289912)(2,0.389268)(3,0.341208)(4,0.348134)(5,0.350717)(6,0.318911)(7,0.321720)(8,0.316741)(9,0.313930)(10,0.312931)(11,0.318187)(12,0.313015)(13,0.313414)(14,0.308711)(15,0.303176)(16,0.303486)(17,0.304351)(18,0.302172)(19,0.302367)(20,0.291263)(21,0.264139)(22,0.248613)(23,0.246281)(24,0.246907)(25,0.244825)
                };
                \addplot[color=blue, mark=square*, solid] coordinates {
                  (1,0.322756)(2,0.355379)(3,0.373688)(4,0.396873)(5,0.411509)(6,0.408096)(7,0.399590)(8,0.389824)(9,0.373282)(10,0.350131)(11,0.319400)(12,0.303736)(13,0.302218)(14,0.294804)(15,0.275604)(16,0.264335)(17,0.256131)(18,0.254999)(19,0.258761)(20,0.256967)(21,0.258673)(22,0.260006)(23,0.257671)(24,0.257272)(25,0.253287)
                };
                \addplot[color=green!50!black, mark=square*, solid] coordinates {
                  (1,0.355784)(2,0.386085)(3,0.375447)(4,0.379824)(5,0.383614)(6,0.381126)(7,0.376388)(8,0.369703)(9,0.360067)(10,0.343818)(11,0.340861)(12,0.333089)(13,0.332481)(14,0.327519)(15,0.325585)(16,0.325118)(17,0.318683)(18,0.315994)(19,0.314385)(20,0.306946)(21,0.308376)(22,0.300432)(23,0.299099)(24,0.295079)(25,0.283971)
                };
                \addplot[color=red, mark=triangle*, solid] coordinates {
                  (1,0.250784)(2,0.274677)(3,0.263630)(4,0.241595)(5,0.217642)(6,0.201324)(7,0.192234)(8,0.188677)(9,0.184232)(10,0.182744)(11,0.179671)(12,0.175853)(13,0.173421)(14,0.171616)(15,0.169083)(16,0.163129)(17,0.159475)(18,0.155928)(19,0.152836)(20,0.154065)(21,0.151076)(22,0.151132)(23,0.148961)(24,0.150713)(25,0.146424)
                };
                \addplot[color=blue, mark=triangle*, solid] coordinates {
                  (1,0.265651)(2,0.247377)(3,0.238435)(4,0.229893)(5,0.224802)(6,0.209586)(7,0.203061)(8,0.193698)(9,0.189999)(10,0.179305)(11,0.172868)(12,0.161986)(13,0.159487)(14,0.149762)(15,0.150659)(16,0.143353)(17,0.144291)(18,0.142187)(19,0.140549)(20,0.138904)(21,0.139245)(22,0.135951)(23,0.136412)(24,0.135147)(25,0.133519)
                };
                \addplot[color=green!50!black, mark=triangle*, solid] coordinates {
                  (1,0.281420)(2,0.270902)(3,0.255228)(4,0.238218)(5,0.222032)(6,0.203676)(7,0.195161)(8,0.184239)(9,0.175642)(10,0.170222)(11,0.160788)(12,0.151992)(13,0.144233)(14,0.142047)(15,0.136989)(16,0.137598)(17,0.135734)(18,0.139074)(19,0.135062)(20,0.136151)(21,0.132686)(22,0.134928)(23,0.129481)(24,0.127994)(25,0.121795)
                };
                \addplot[color=red, mark=diamond*, solid] coordinates {
                  (1,0.236338)(2,0.232545)(3,0.220356)(4,0.209022)(5,0.198619)(6,0.191803)(7,0.183723)(8,0.178957)(9,0.172166)(10,0.169724)(11,0.164211)(12,0.161159)(13,0.155907)(14,0.153645)(15,0.145727)(16,0.144173)(17,0.138937)(18,0.139376)(19,0.134648)(20,0.133759)(21,0.129308)(22,0.129802)(23,0.127225)(24,0.127659)(25,0.125488)
                };
                \addplot[color=blue, mark=diamond*, solid] coordinates {
                  (1,0.244535)(2,0.221794)(3,0.201298)(4,0.184062)(5,0.174849)(6,0.164118)(7,0.157911)(8,0.151682)(9,0.148282)(10,0.144389)(11,0.140109)(12,0.135252)(13,0.131600)(14,0.129846)(15,0.125993)(16,0.124803)(17,0.122303)(18,0.119673)(19,0.119247)(20,0.118408)(21,0.116958)(22,0.113251)(23,0.113424)(24,0.111852)(25,0.112244)
                };
                \addplot[color=green!50!black, mark=diamond*, solid] coordinates {
                  (1,0.222455)(2,0.199806)(3,0.185153)(4,0.165700)(5,0.154487)(6,0.141683)(7,0.135559)(8,0.136108)(9,0.128187)(10,0.125216)(11,0.121309)(12,0.121266)(13,0.115162)(14,0.112487)(15,0.109623)(16,0.107709)(17,0.102933)(18,0.102299)(19,0.099385)(20,0.098896)(21,0.096985)(22,0.096313)(23,0.094520)(24,0.094208)(25,0.092229)
                };
            \end{axis}
        \end{tikzpicture}
        
    \end{minipage}}\hspace{-4em}%
    \subfloat{
        \begin{minipage}[t]{0.49\textwidth}
        \centering
        \begin{tikzpicture}[trim axis left, trim axis right]
            \begin{axis}[
                 width=\textwidth,
                 height=0.85\textwidth,
                 xlabel={Epoch},
                 xtick={0,5,10,15,20,25},
                 xmin=-1.5, xmax=26.5,
                 ylabel={},
                 grid=both,
                 grid style={line width=.1pt, dashed, draw=gray!10},
                 major grid style={line width=.2pt,draw=gray!50},
                 label style={font=\scriptsize},
                 tick label style={font=\scriptsize},
                 every axis title/.style={at={(0.5,1.0)}, above, font=\small},
                 title={CLIP: BatchTopKSAE},
                 mark size=1.2pt,
                 line width=0.6pt,
                 scaled y ticks=false,
                 yticklabel style={
                     /pgf/number format/fixed,
                     /pgf/number format/fixed zerofill,
                     /pgf/number format/precision=2
                 },
                 axis y line*=right,
                 axis x line*=bottom,
                 mark size=0.8pt,
                 line width=0.4pt,
                 ytick={0.05, 0.1, 0.15,0.2,0.25,0.3,0.35,0.4,0.45,0.5},
                 ymin=0.025,
                 ]
                \addplot[color=red, mark=square*, solid] coordinates {
                  (1,0.394692)(2,0.182103)(3,0.167368)(4,0.133531)(5,0.122689)(6,0.129550)(7,0.146349)(8,0.161402)(9,0.173568)(10,0.174239)(11,0.166913)(12,0.150742)(13,0.123779)(14,0.115260)(15,0.110775)(16,0.116148)(17,0.123400)(18,0.133572)(19,0.145464)(20,0.146504)(21,0.148083)(22,0.137828)(23,0.128477)(24,0.128916)(25,0.133802)
                };
                \addplot[color=blue, mark=square*, solid] coordinates {
                  (1,0.281051)(2,0.143315)(3,0.169979)(4,0.184175)(5,0.164109)(6,0.146594)(7,0.136849)(8,0.140097)(9,0.145129)(10,0.141738)(11,0.121666)(12,0.101441)(13,0.094217)(14,0.091410)(15,0.094690)(16,0.100021)(17,0.106050)(18,0.112966)(19,0.115696)(20,0.120623)(21,0.132033)(22,0.138023)(23,0.139127)(24,0.142712)(25,0.150898)
                };
                \addplot[color=green!50!black, mark=square*, solid] coordinates {
                  (1,0.227555)(2,0.145162)(3,0.132442)(4,0.109292)(5,0.087921)(6,0.097507)(7,0.134859)(8,0.161460)(9,0.145377)(10,0.133255)(11,0.133394)(12,0.134295)(13,0.133936)(14,0.129545)(15,0.119574)(16,0.103050)(17,0.093197)(18,0.089564)(19,0.085567)(20,0.081753)(21,0.080754)(22,0.078306)(23,0.079228)(24,0.080775)(25,0.081924)
                };
                \addplot[color=red, mark=triangle*, solid] coordinates {
                  (1,0.339161)(2,0.191878)(3,0.164228)(4,0.161232)(5,0.154052)(6,0.143130)(7,0.133235)(8,0.120680)(9,0.114553)(10,0.110256)(11,0.110716)(12,0.109751)(13,0.115659)(14,0.115520)(15,0.118999)(16,0.109612)(17,0.111182)(18,0.110201)(19,0.116339)(20,0.119647)(21,0.122002)(22,0.121821)(23,0.129424)(24,0.124864)(25,0.124194)
                };
                \addplot[color=blue, mark=triangle*, solid] coordinates {
                  (1,0.279237)(2,0.163367)(3,0.139813)(4,0.127084)(5,0.111079)(6,0.100988)(7,0.092237)(8,0.085707)(9,0.087102)(10,0.084564)(11,0.084879)(12,0.083696)(13,0.083424)(14,0.085693)(15,0.089428)(16,0.084523)(17,0.087577)(18,0.082627)(19,0.083095)(20,0.082865)(21,0.087244)(22,0.087472)(23,0.092812)(24,0.094823)(25,0.098987)
                };
                \addplot[color=green!50!black, mark=triangle*, solid] coordinates {
                  (1,0.290286)(2,0.155301)(3,0.110343)(4,0.100718)(5,0.089742)(6,0.082692)(7,0.075634)(8,0.069985)(9,0.065884)(10,0.063924)(11,0.061503)(12,0.062079)(13,0.061223)(14,0.063412)(15,0.063699)(16,0.064895)(17,0.063644)(18,0.065543)(19,0.065544)(20,0.065871)(21,0.065147)(22,0.067788)(23,0.067666)(24,0.067691)(25,0.068434)
                };
                \addplot[color=red, mark=diamond*, solid] coordinates {
                  (1,0.254381)(2,0.203036)(3,0.201530)(4,0.184747)(5,0.179810)(6,0.167587)(7,0.160590)(8,0.155014)(9,0.150817)(10,0.146236)(11,0.142328)(12,0.138271)(13,0.136973)(14,0.134180)(15,0.133029)(16,0.129621)(17,0.127009)(18,0.122777)(19,0.122865)(20,0.119283)(21,0.119030)(22,0.116307)(23,0.115104)(24,0.113243)(25,0.112507)
                };
                \addplot[color=blue, mark=diamond*, solid] coordinates {
                  (1,0.226034)(2,0.165107)(3,0.158402)(4,0.145061)(5,0.133833)(6,0.125853)(7,0.119203)(8,0.116638)(9,0.112612)(10,0.109930)(11,0.106791)(12,0.103599)(13,0.099717)(14,0.100380)(15,0.097076)(16,0.096333)(17,0.094559)(18,0.092373)(19,0.091767)(20,0.091253)(21,0.089443)(22,0.089572)(23,0.088768)(24,0.087318)(25,0.087572)
                };
                \addplot[color=green!50!black, mark=diamond*, solid] coordinates {
                  (1,0.227825)(2,0.150121)(3,0.134425)(4,0.118374)(5,0.107166)(6,0.099518)(7,0.093962)(8,0.089654)(9,0.086162)(10,0.083786)(11,0.081356)(12,0.079070)(13,0.075108)(14,0.073921)(15,0.072512)(16,0.070160)(17,0.068579)(18,0.067229)(19,0.067064)(20,0.065653)(21,0.065195)(22,0.065057)(23,0.064678)(24,0.065218)(25,0.064200)
                };
            \end{axis}
        \end{tikzpicture}
    \end{minipage}}\hfill%

    \vspace{0.4em}

    \subfloat{
        \begin{minipage}[t]{0.49\textwidth}
        \centering
        \begin{tikzpicture}[trim axis left, trim axis right]
            \begin{axis}[
                 width=\textwidth,
                 height=0.85\textwidth,
                 xlabel={Epoch},
                 xtick={0,5,10,15,20,25},
                 xmin=-1.5, xmax=26.5,
                 ylabel={TMS},
                 grid=both,
                 grid style={line width=.1pt, dashed, draw=gray!10},
                 major grid style={line width=.2pt,draw=gray!50},
                 label style={font=\scriptsize},
                 tick label style={font=\scriptsize},
                 every axis title/.style={at={(0.5,1.0)}, above, font=\small},
                 title={BLIP: TopKSAE},
                 mark size=1.2pt,
                 line width=0.6pt,
                 scaled y ticks=false,
                 yticklabel style={
                     /pgf/number format/fixed,
                     /pgf/number format/fixed zerofill,
                     /pgf/number format/precision=2
                 },
                 axis y line*=left,
                 axis x line*=bottom,
                 mark size=0.8pt,
                 line width=0.4pt,
                 ytick={0.05, 0.1, 0.15,0.2,0.25,0.3,0.35,0.4,0.45,0.5},
                 ]
                \addplot[color=red, mark=square*, solid] coordinates {
                  (1,0.323826)(2,0.365516)(3,0.362453)(4,0.331551)(5,0.321685)(6,0.314999)(7,0.300334)(8,0.287384)(9,0.277604)(10,0.266945)(11,0.270444)(12,0.258107)(13,0.256685)(14,0.258665)(15,0.247954)(16,0.244854)(17,0.247563)(18,0.242405)(19,0.230134)(20,0.221272)(21,0.211479)(22,0.211874)(23,0.228547)(24,0.210232)(25,0.209844)
                };
                \addplot[color=blue, mark=square*, solid] coordinates {
                  (1,0.397021)(2,0.386579)(3,0.387825)(4,0.364723)(5,0.359187)(6,0.338562)(7,0.338520)(8,0.298890)(9,0.286864)(10,0.290853)(11,0.280945)(12,0.272132)(13,0.269626)(14,0.268175)(15,0.258681)(16,0.259181)(17,0.261886)(18,0.257573)(19,0.251436)(20,0.248426)(21,0.239353)(22,0.229396)(23,0.219100)(24,0.218245)(25,0.221467)
                };
                \addplot[color=green!50!black, mark=square*, solid] coordinates {
                  (1,0.376369)(2,0.423201)(3,0.464575)(4,0.446655)(5,0.440018)(6,0.417797)(7,0.420161)(8,0.371095)(9,0.353628)(10,0.352692)(11,0.348442)(12,0.330378)(13,0.318526)(14,0.291211)(15,0.288583)(16,0.270675)(17,0.276713)(18,0.268984)(19,0.265391)(20,0.269639)(21,0.258413)(22,0.258215)(23,0.250770)(24,0.245399)(25,0.233833)
                };
                \addplot[color=red, mark=triangle*, solid] coordinates {
                  (1,0.273851)(2,0.296461)(3,0.249239)(4,0.236901)(5,0.210954)(6,0.183868)(7,0.183766)(8,0.175561)(9,0.172704)(10,0.161458)(11,0.163045)(12,0.152097)(13,0.149990)(14,0.153176)(15,0.149427)(16,0.145713)(17,0.147732)(18,0.142948)(19,0.144456)(20,0.136372)(21,0.142269)(22,0.134794)(23,0.138406)(24,0.129169)(25,0.133129)
                };
                \addplot[color=blue, mark=triangle*, solid] coordinates {
                  (1,0.305592)(2,0.300906)(3,0.234721)(4,0.255255)(5,0.218812)(6,0.206607)(7,0.182781)(8,0.191588)(9,0.184748)(10,0.172810)(11,0.173896)(12,0.158937)(13,0.163893)(14,0.152116)(15,0.150340)(16,0.146060)(17,0.148446)(18,0.138273)(19,0.141070)(20,0.132356)(21,0.131339)(22,0.127489)(23,0.125759)(24,0.129067)(25,0.124474)
                };
                \addplot[color=green!50!black, mark=triangle*, solid] coordinates {
                  (1,0.315708)(2,0.319155)(3,0.244027)(4,0.257779)(5,0.221006)(6,0.217902)(7,0.191937)(8,0.212705)(9,0.182656)(10,0.202551)(11,0.178704)(12,0.177177)(13,0.171518)(14,0.174962)(15,0.161847)(16,0.167041)(17,0.158255)(18,0.154386)(19,0.147622)(20,0.146082)(21,0.136095)(22,0.136660)(23,0.129513)(24,0.131357)(25,0.132447)
                };
                \addplot[color=red, mark=diamond*, solid] coordinates {
                  (1,0.261415)(2,0.245129)(3,0.201819)(4,0.225578)(5,0.185223)(6,0.178546)(7,0.183772)(8,0.179487)(9,0.170779)(10,0.179694)(11,0.162946)(12,0.173032)(13,0.162249)(14,0.164417)(15,0.150597)(16,0.156047)(17,0.142755)(18,0.142835)(19,0.139309)(20,0.147646)(21,0.136949)(22,0.149077)(23,0.135659)(24,0.129926)(25,0.135217)
                };
                \addplot[color=blue, mark=diamond*, solid] coordinates {
                  (1,0.277242)(2,0.262617)(3,0.175535)(4,0.216704)(5,0.164304)(6,0.179681)(7,0.175610)(8,0.195097)(9,0.156380)(10,0.170942)(11,0.161913)(12,0.161239)(13,0.161035)(14,0.155805)(15,0.163859)(16,0.147481)(17,0.148374)(18,0.141782)(19,0.147516)(20,0.133595)(21,0.139332)(22,0.131254)(23,0.136715)(24,0.125542)(25,0.131153)
                };
                \addplot[color=green!50!black, mark=diamond*, solid] coordinates {
                  (1,0.289153)(2,0.287412)(3,0.200048)(4,0.220459)(5,0.176908)(6,0.188080)(7,0.178267)(8,0.189114)(9,0.171045)(10,0.180197)(11,0.166611)(12,0.190359)(13,0.168361)(14,0.171406)(15,0.161953)(16,0.176624)(17,0.145093)(18,0.163990)(19,0.148208)(20,0.155018)(21,0.141092)(22,0.157418)(23,0.132054)(24,0.144069)(25,0.139049)
                };
            \end{axis}
        \end{tikzpicture}
    \end{minipage}}\hspace{-4em}%
    \subfloat{
        \begin{minipage}[t]{0.49\textwidth}
        \centering
        \begin{tikzpicture}[trim axis left, trim axis right]
            \begin{axis}[
                 width=\textwidth,
                 height=0.85\textwidth,
                 xlabel={Epoch},
                 xtick={0,5,10,15,20,25},
                 xmin=-1.5, xmax=26.5,
                 ylabel={},
                 grid=both,
                 grid style={line width=.1pt, dashed, draw=gray!10},
                 major grid style={line width=.2pt,draw=gray!50},
                 label style={font=\scriptsize},
                 tick label style={font=\scriptsize},
                 every axis title/.style={at={(0.5,1.0)}, above, font=\small},
                 title={BLIP: BatchTopKSAE},
                 mark size=1.2pt,
                 line width=0.6pt,
                 scaled y ticks=false,
                 yticklabel style={
                     /pgf/number format/fixed,
                     /pgf/number format/fixed zerofill,
                     /pgf/number format/precision=2
                 },
                 axis y line*=right,
                 axis x line*=bottom,
                 mark size=0.8pt,
                 line width=0.4pt,
                 ytick={0.05, 0.1, 0.15,0.2,0.25,0.3,0.35,0.4,0.45,0.5},
                 ]
                \addplot[color=red, mark=square*, solid] coordinates {
                  (1,0.197207)(2,0.182452)(3,0.199186)(4,0.195231)(5,0.168435)(6,0.139878)(7,0.115138)(8,0.101548)(9,0.088197)(10,0.080644)(11,0.082070)(12,0.086033)(13,0.086913)(14,0.081485)(15,0.088519)(16,0.091442)(17,0.090309)(18,0.090019)(19,0.094204)(20,0.101899)(21,0.103084)(22,0.109666)(23,0.107637)(24,0.112690)(25,0.115475)
                };
                \addplot[color=blue, mark=square*, solid] coordinates {
                  (1,0.169197)(2,0.150445)(3,0.175901)(4,0.168177)(5,0.154721)(6,0.138567)(7,0.134609)(8,0.135499)(9,0.134579)(10,0.130921)(11,0.127179)(12,0.118661)(13,0.117271)(14,0.109154)(15,0.106677)(16,0.103682)(17,0.104182)(18,0.103594)(19,0.104091)(20,0.113024)(21,0.115083)(22,0.119386)(23,0.112854)(24,0.106317)(25,0.100793)
                };
                \addplot[color=green!50!black, mark=square*, solid] coordinates {
                  (1,0.173953)(2,0.133762)(3,0.121689)(4,0.119358)(5,0.119157)(6,0.111911)(7,0.110934)(8,0.100991)(9,0.102404)(10,0.098449)(11,0.099242)(12,0.099028)(13,0.094835)(14,0.089376)(15,0.082440)(16,0.078318)(17,0.078353)(18,0.076887)(19,0.088669)(20,0.093785)(21,0.102106)(22,0.108618)(23,0.121110)(24,0.127876)(25,0.133645)
                };
                \addplot[color=red, mark=triangle*, solid] coordinates {
                  (1,0.217381)(2,0.201583)(3,0.162826)(4,0.124894)(5,0.107193)(6,0.108673)(7,0.099787)(8,0.094172)(9,0.094143)(10,0.091946)(11,0.093613)(12,0.090560)(13,0.094675)(14,0.087763)(15,0.086244)(16,0.086785)(17,0.084392)(18,0.077779)(19,0.079386)(20,0.075569)(21,0.074856)(22,0.075716)(23,0.075231)(24,0.073670)(25,0.073461)
                };
                \addplot[color=blue, mark=triangle*, solid] coordinates {
                  (1,0.200935)(2,0.174038)(3,0.145250)(4,0.131546)(5,0.094466)(6,0.091470)(7,0.078516)(8,0.078761)(9,0.068997)(10,0.068524)(11,0.063850)(12,0.064509)(13,0.060522)(14,0.060953)(15,0.061080)(16,0.059830)(17,0.058564)(18,0.057010)(19,0.055110)(20,0.054686)(21,0.054200)(22,0.053796)(23,0.053727)(24,0.053715)(25,0.053330)
                };
                \addplot[color=green!50!black, mark=triangle*, solid] coordinates {
                  (1,0.180562)(2,0.161438)(3,0.144528)(4,0.119228)(5,0.108854)(6,0.084470)(7,0.078291)(8,0.062549)(9,0.065447)(10,0.059631)(11,0.061242)(12,0.058095)(13,0.056946)(14,0.056979)(15,0.053156)(16,0.049161)(17,0.046086)(18,0.044380)(19,0.043612)(20,0.043262)(21,0.042722)(22,0.042383)(23,0.042667)(24,0.040999)(25,0.040917)
                };
                \addplot[color=red, mark=diamond*, solid] coordinates {
                  (1,0.236562)(2,0.190835)(3,0.167152)(4,0.142950)(5,0.158939)(6,0.136297)(7,0.139029)(8,0.133886)(9,0.128538)(10,0.130608)(11,0.131441)(12,0.121529)(13,0.124604)(14,0.111470)(15,0.111059)(16,0.107145)(17,0.101808)(18,0.098209)(19,0.100543)(20,0.096640)(21,0.096993)(22,0.092979)(23,0.094599)(24,0.091730)(25,0.090074)
                };
                \addplot[color=blue, mark=diamond*, solid] coordinates {
                  (1,0.233246)(2,0.179740)(3,0.152555)(4,0.129390)(5,0.134776)(6,0.109082)(7,0.144899)(8,0.104932)(9,0.117621)(10,0.112534)(11,0.103465)(12,0.109506)(13,0.096888)(14,0.098324)(15,0.094562)(16,0.091180)(17,0.083847)(18,0.089111)(19,0.083452)(20,0.081261)(21,0.080382)(22,0.077960)(23,0.074471)(24,0.074660)(25,0.074480)
                };
                \addplot[color=green!50!black, mark=diamond*, solid] coordinates {
                  (1,0.250264)(2,0.184233)(3,0.147711)(4,0.120947)(5,0.111269)(6,0.101209)(7,0.119189)(8,0.094497)(9,0.108821)(10,0.091401)(11,0.102566)(12,0.086431)(13,0.096134)(14,0.081994)(15,0.084348)(16,0.080296)(17,0.079827)(18,0.078061)(19,0.071792)(20,0.073900)(21,0.069378)(22,0.071606)(23,0.064664)(24,0.073549)(25,0.062545)
                };
            \end{axis}
        \end{tikzpicture}
    \end{minipage}}

    \vspace{0.3em}
    \ref{legendTopK}\par\vspace{0.2em}
    \ref{legendBatch}

    \caption{Epoch evolution of TMS for TopKSAE (left column) and BatchTopKSAE (right column) across DINO, CLIP, and BLIP models.}
    \label{fig:epoch_evolution_tversky_similarity}
\end{figure*}

%% file: FIGS/corr/target_drop_combined_plot.tex
\begin{figure*}[h]
    \centering
    \subfloat{
        \begin{minipage}[t]{0.49\textwidth}
        \centering
        \begin{tikzpicture}[trim axis left, trim axis right]
            \begin{axis}[
                 width=\textwidth,
                 height=0.75\textwidth,
                 xlabel={Number of deleted features (N)},
                 ylabel={Target drop},
                 grid=both,
                 grid style={line width=.1pt, dashed, draw=gray!10},
                 major grid style={line width=.2pt,draw=gray!50},
                 xmode=log,
                 log basis x=2,
                 xtick={1,2,4,8,16,32},
                 xticklabels={$2^0$,$2^1$,$2^2$,$2^3$,$2^4$,$2^5$},
                 ytick={0.0, 0.1, 0.2, 0.3, 0.4, 0.5, 0.6, 0.7, 0.8, 0.9, 1.0},
                 yticklabel style={
                     /pgf/number format/fixed,
                     /pgf/number format/fixed zerofill,
                     /pgf/number format/precision=1
                 },
                 label style={font=\scriptsize},
                 tick label style={font=\scriptsize},
                 every axis title/.style={at={(0.5,1.0)}, above, font=\small},
                 title={DINO: TopKSAE},
                 mark size=0.8pt,
                 line width=0.4pt,
                 legend to name=legendTopK,
                 legend style={legend columns=3, font=\scriptsize, column sep=0.3cm},
                 axis x line*=bottom,
                 axis y line*=left,
                 ]
                \addplot[color=red, mark=square*, solid] coordinates {
                  (1,0.06451)(2,0.14111)(4,0.29186)(8,0.60043)(16,0.8212)(32,0.8898)
                };
                \addlegendentry{Exp=2, K=10}
                \addplot[color=blue, mark=square*, solid] coordinates {
                  (1,0.04897)(2,0.10674)(4,0.22006)(8,0.50889)(16,0.75246)(32,0.8716)
                };
                \addlegendentry{Exp=4, K=10}
                \addplot[color=green!50!black, mark=square*, solid] coordinates {
                  (1,0.04491)(2,0.09769)(4,0.21774)(8,0.45734)(16,0.70703)(32,0.84709)
                };
                \addlegendentry{Exp=8, K=10}
                \addplot[color=red, mark=triangle*, solid] coordinates {
                  (1,0.02997)(2,0.05989)(4,0.14037)(8,0.29374)(16,0.57934)(32,0.84246)
                };
                \addlegendentry{Exp=2, K=32}
                \addplot[color=blue, mark=triangle*, solid] coordinates {
                  (1,0.02)(2,0.04757)(4,0.09557)(8,0.20814)(16,0.49254)(32,0.76623)
                };
                \addlegendentry{Exp=4, K=32}
                \addplot[color=green!50!black, mark=triangle*, solid] coordinates {
                  (1,0.0186)(2,0.0366)(4,0.07686)(8,0.18606)(16,0.434)(32,0.71203)
                };
                \addlegendentry{Exp=8, K=32}
                \addplot[color=red, mark=diamond*, solid] coordinates {
                  (1,0.02294)(2,0.04571)(4,0.09377)(8,0.19506)(16,0.39689)(32,0.68403)
                };
                \addlegendentry{Exp=2, K=64}
                \addplot[color=blue, mark=diamond*, solid] coordinates {
                  (1,0.01589)(2,0.03234)(4,0.07137)(8,0.15106)(16,0.31566)(32,0.62643)
                };
                \addlegendentry{Exp=4, K=64}
                \addplot[color=green!50!black, mark=diamond*, solid] coordinates {
                  (1,0.01174)(2,0.02269)(4,0.0514)(8,0.112)(16,0.25637)(32,0.5194)
                };
                \addlegendentry{Exp=8, K=64}
            \end{axis}
        \end{tikzpicture}
    \end{minipage}}\hspace{-4em}%
    \subfloat{
        \begin{minipage}[t]{0.49\textwidth}
        \centering
        \begin{tikzpicture}[trim axis left, trim axis right]
            \begin{axis}[
                 width=\textwidth,
                 height=0.75\textwidth,
                 xlabel={Number of deleted features (N)},
                 ylabel={},
                 grid=both,
                 grid style={line width=.1pt, dashed, draw=gray!10},
                 major grid style={line width=.2pt,draw=gray!50},
                 xmode=log,
                 log basis x=2,
                 xtick={1,2,4,8,16,32},
                 xticklabels={$2^0$,$2^1$,$2^2$,$2^3$,$2^4$,$2^5$},
                 ytick={0.0, 0.1, 0.2, 0.3, 0.4, 0.5, 0.6, 0.7, 0.8, 0.9},
                 yticklabel style={
                     /pgf/number format/fixed,
                     /pgf/number format/fixed zerofill,
                     /pgf/number format/precision=1
                 },
                 label style={font=\scriptsize},
                 tick label style={font=\scriptsize},
                 every axis title/.style={at={(0.5,1.0)}, above, font=\small},
                 title={DINO: BatchTopKSAE},
                 mark size=0.8pt,
                 line width=0.4pt,
                 legend to name=legendBatch,
                 legend style={legend columns=3, font=\scriptsize, column sep=0.3cm},
                 axis x line*=bottom,
                 axis y line*=right,
                 ]
                \addplot[color=red, mark=square*, solid] coordinates {
                  (1,0.04969)(2,0.10334)(4,0.2442)(8,0.58203)(16,0.79509)(32,0.85169)
                };
                \addplot[color=blue, mark=square*, solid] coordinates {
                  (1,0.02974)(2,0.07529)(4,0.17831)(8,0.4444)(16,0.70034)(32,0.78763)
                };
                \addplot[color=green!50!black, mark=square*, solid] coordinates {
                  (1,0.03249)(2,0.06569)(4,0.14463)(8,0.40517)(16,0.65169)(32,0.76911)
                };
                \addplot[color=red, mark=triangle*, solid] coordinates {
                  (1,0.03109)(2,0.07129)(4,0.15334)(8,0.28643)(16,0.53686)(32,0.8164)
                };
                \addplot[color=blue, mark=triangle*, solid] coordinates {
                  (1,0.02234)(2,0.04154)(4,0.09191)(8,0.20177)(16,0.44834)(32,0.74977)
                };
                \addplot[color=green!50!black, mark=triangle*, solid] coordinates {
                  (1,0.01534)(2,0.03163)(4,0.06823)(8,0.1514)(16,0.3694)(32,0.65477)
                };
                \addplot[color=red, mark=diamond*, solid] coordinates {
                  (1,0.02083)(2,0.045)(4,0.09837)(8,0.20271)(16,0.40337)(32,0.70714)
                };
                \addplot[color=blue, mark=diamond*, solid] coordinates {
                  (1,0.0156)(2,0.03146)(4,0.06583)(8,0.12717)(16,0.30046)(32,0.6092)
                };
                \addplot[color=green!50!black, mark=diamond*, solid] coordinates {
                  (1,0.01186)(2,0.02414)(4,0.04937)(8,0.10357)(16,0.25349)(32,0.52974)
                };
            \end{axis}
        \end{tikzpicture}
    \end{minipage}}\hfill%

    \vspace{0.4em}

    \subfloat{
        \begin{minipage}[t]{0.49\textwidth}
        \centering
        \begin{tikzpicture}[trim axis left, trim axis right]
            \begin{axis}[
                 width=\textwidth,
                 height=0.75\textwidth,
                 xlabel={Number of deleted features (N)},
                 ylabel={Target drop},
                 grid=both,
                 grid style={line width=.1pt, dashed, draw=gray!10},
                 major grid style={line width=.2pt,draw=gray!50},
                 xmode=log,
                 log basis x=2,
                 xtick={1,2,4,8,16,32},
                 xticklabels={$2^0$,$2^1$,$2^2$,$2^3$,$2^4$,$2^5$},
                 ytick={0.0, 0.1, 0.2, 0.3, 0.4, 0.5, 0.6, 0.7, 0.8, 0.9},
                 yticklabel style={
                     /pgf/number format/fixed,
                     /pgf/number format/fixed zerofill,
                     /pgf/number format/precision=1
                 },
                 label style={font=\scriptsize},
                 tick label style={font=\scriptsize},
                 every axis title/.style={at={(0.5,1.0)}, above, font=\small},
                 title={CLIP: TopKSAE},
                 mark size=0.8pt,
                 line width=0.4pt,
                 axis x line*=bottom,
                 axis y line*=left,
                 ]
                \addplot[color=red, mark=square*, solid] coordinates {
                  (1,0.47)(2,0.575)(4,0.642)(8,0.668)(16,0.678)(32,0.682)
                };
                \addplot[color=blue, mark=square*, solid] coordinates {
                  (1,0.459)(2,0.572)(4,0.639)(8,0.671)(16,0.683)(32,0.688)
                };
                \addplot[color=green!50!black, mark=square*, solid] coordinates {
                  (1,0.423)(2,0.567)(4,0.658)(8,0.7)(16,0.713)(32,0.718)
                };
                \addplot[color=red, mark=triangle*, solid] coordinates {
                  (1,0.207)(2,0.383)(4,0.603)(8,0.765)(16,0.829)(32,0.849)
                };
                \addplot[color=blue, mark=triangle*, solid] coordinates {
                  (1,0.184)(2,0.332)(4,0.526)(8,0.706)(16,0.792)(32,0.83)
                };
                \addplot[color=green!50!black, mark=triangle*, solid] coordinates {
                  (1,0.138)(2,0.265)(4,0.456)(8,0.639)(16,0.741)(32,0.799)
                };
                \addplot[color=red, mark=diamond*, solid] coordinates {
                  (1,0.101)(2,0.199)(4,0.379)(8,0.611)(16,0.781)(32,0.858)
                };
                \addplot[color=blue, mark=diamond*, solid] coordinates {
                  (1,0.068)(2,0.132)(4,0.249)(8,0.457)(16,0.65)(32,0.799)
                };
                \addplot[color=green!50!black, mark=diamond*, solid] coordinates {
                  (1,0.056)(2,0.102)(4,0.191)(8,0.34)(16,0.531)(32,0.695)
                };
            \end{axis}
        \end{tikzpicture}
        
    \end{minipage}}\hspace{-4em}%
    \subfloat{
        \begin{minipage}[t]{0.49\textwidth}
        \centering
        \begin{tikzpicture}[trim axis left, trim axis right]
            \begin{axis}[
                 width=\textwidth,
                 height=0.75\textwidth,
                 xlabel={Number of deleted features (N)},
                 ylabel={},
                 grid=both,
                 grid style={line width=.1pt, dashed, draw=gray!10},
                 major grid style={line width=.2pt,draw=gray!50},
                 xmode=log,
                 log basis x=2,
                 xtick={1,2,4,8,16,32},
                 xticklabels={$2^0$,$2^1$,$2^2$,$2^3$,$2^4$,$2^5$},
                 ytick={0.0, 0.1, 0.2, 0.3, 0.4, 0.5, 0.6, 0.7, 0.8, 0.9},
                 yticklabel style={
                     /pgf/number format/fixed,
                     /pgf/number format/fixed zerofill,
                     /pgf/number format/precision=1
                 },
                 label style={font=\scriptsize},
                 tick label style={font=\scriptsize},
                 every axis title/.style={at={(0.5,1.0)}, above, font=\small},
                 title={CLIP: BatchTopKSAE},
                 mark size=0.8pt,
                 line width=0.4pt,
                 axis x line*=bottom,
                 axis y line*=right,
                 ]
                \addplot[color=red, mark=square*, solid] coordinates {
                  (1,0.342)(2,0.415)(4,0.474)(8,0.529)(16,0.592)(32,0.666)
                };
                \addplot[color=blue, mark=square*, solid] coordinates {
                  (1,0.298)(2,0.353)(4,0.395)(8,0.431)(16,0.463)(32,0.493)
                };
                \addplot[color=green!50!black, mark=square*, solid] coordinates {
                  (1,0.261)(2,0.304)(4,0.326)(8,0.348)(16,0.374)(32,0.404)
                };
                \addplot[color=red, mark=triangle*, solid] coordinates {
                  (1,0.202)(2,0.371)(4,0.587)(8,0.726)(16,0.806)(32,0.853)
                };
                \addplot[color=blue, mark=triangle*, solid] coordinates {
                  (1,0.154)(2,0.276)(4,0.456)(8,0.597)(16,0.696)(32,0.769)
                };
                \addplot[color=green!50!black, mark=triangle*, solid] coordinates {
                  (1,0.106)(2,0.206)(4,0.37)(8,0.509)(16,0.595)(32,0.663)
                };
                \addplot[color=red, mark=diamond*, solid] coordinates {
                  (1,0.097)(2,0.193)(4,0.371)(8,0.605)(16,0.77)(32,0.857)
                };
                \addplot[color=blue, mark=diamond*, solid] coordinates {
                  (1,0.064)(2,0.13)(4,0.241)(8,0.445)(16,0.643)(32,0.795)
                };
                \addplot[color=green!50!black, mark=diamond*, solid] coordinates {
                  (1,0.041)(2,0.084)(4,0.171)(8,0.301)(16,0.474)(32,0.639)
                };
            \end{axis}
        \end{tikzpicture}
    \end{minipage}}\hfill%

    \vspace{0.4em}

    \subfloat{
        \begin{minipage}[t]{0.49\textwidth}
        \centering
        \begin{tikzpicture}[trim axis left, trim axis right]
            \begin{axis}[
                 width=\textwidth,
                 height=0.75\textwidth,
                 xlabel={Number of deleted features (N)},
                 ylabel={Target drop},
                 grid=both,
                 grid style={line width=.1pt, dashed, draw=gray!10},
                 major grid style={line width=.2pt,draw=gray!50},
                 xmode=log,
                 log basis x=2,
                 xtick={1,2,4,8,16,32},
                 xticklabels={$2^0$,$2^1$,$2^2$,$2^3$,$2^4$,$2^5$},
                 ytick={0.0, 0.1, 0.2, 0.3, 0.4, 0.5, 0.6, 0.7, 0.8, 0.9},
                 yticklabel style={
                     /pgf/number format/fixed,
                     /pgf/number format/fixed zerofill,
                     /pgf/number format/precision=1
                 },
                 label style={font=\scriptsize},
                 tick label style={font=\scriptsize},
                 every axis title/.style={at={(0.5,1.0)}, above, font=\small},
                 title={BLIP: TopKSAE},
                 mark size=0.8pt,
                 line width=0.4pt,
                 axis x line*=bottom,
                 axis y line*=left,
                 ]
                \addplot[color=red, mark=square*, solid] coordinates {
                  (1,0.486)(2,0.68)(4,0.782)(8,0.847)(16,0.878)(32,0.89)
                };
                \addplot[color=blue, mark=square*, solid] coordinates {
                  (1,0.511)(2,0.71)(4,0.815)(8,0.864)(16,0.889)(32,0.901)
                };
                \addplot[color=green!50!black, mark=square*, solid] coordinates {
                  (1,0.493)(2,0.671)(4,0.818)(8,0.877)(16,0.902)(32,0.907)
                };
                \addplot[color=red, mark=triangle*, solid] coordinates {
                  (1,0.09)(2,0.192)(4,0.384)(8,0.605)(16,0.787)(32,0.895)
                };
                \addplot[color=blue, mark=triangle*, solid] coordinates {
                  (1,0.067)(2,0.153)(4,0.299)(8,0.489)(16,0.699)(32,0.86)
                };
                \addplot[color=green!50!black, mark=triangle*, solid] coordinates {
                  (1,0.061)(2,0.126)(4,0.251)(8,0.429)(16,0.638)(32,0.829)
                };
                \addplot[color=red, mark=diamond*, solid] coordinates {
                  (1,0.041)(2,0.089)(4,0.181)(8,0.336)(16,0.56)(32,0.789)
                };
                \addplot[color=blue, mark=diamond*, solid] coordinates {
                  (1,0.028)(2,0.069)(4,0.139)(8,0.271)(16,0.465)(32,0.684)
                };
                \addplot[color=green!50!black, mark=diamond*, solid] coordinates {
                  (1,0.033)(2,0.061)(4,0.113)(8,0.218)(16,0.369)(32,0.592)
                };
            \end{axis}
        \end{tikzpicture}
    \end{minipage}}\hspace{-4em}%
    \subfloat{
        \begin{minipage}[t]{0.49\textwidth}
        \centering
        \begin{tikzpicture}[trim axis left, trim axis right]
            \begin{axis}[
                 width=\textwidth,
                 height=0.75\textwidth,
                 xlabel={Number of deleted features (N)},
                 ylabel={},
                 grid=both,
                 grid style={line width=.1pt, dashed, draw=gray!10},
                 major grid style={line width=.2pt,draw=gray!50},
                 xmode=log,
                 log basis x=2,
                 xtick={1,2,4,8,16,32},
                 xticklabels={$2^0$,$2^1$,$2^2$,$2^3$,$2^4$,$2^5$},
                 ytick={0.0, 0.1, 0.2, 0.3, 0.4, 0.5, 0.6, 0.7, 0.8, 0.9},
                 yticklabel style={
                     /pgf/number format/fixed,
                     /pgf/number format/fixed zerofill,
                     /pgf/number format/precision=1
                 },
                 label style={font=\scriptsize},
                 tick label style={font=\scriptsize},
                 every axis title/.style={at={(0.5,1.0)}, above, font=\small},
                 title={BLIP: BatchTopKSAE},
                 mark size=0.8pt,
                 line width=0.4pt,
                 axis x line*=bottom,
                 axis y line*=right,
                 ]
                \addplot[color=red, mark=square*, solid] coordinates {
                  (1,0.286)(2,0.334)(4,0.37)(8,0.42)(16,0.485)(32,0.581)
                };
                \addplot[color=blue, mark=square*, solid] coordinates {
                  (1,0.256)(2,0.376)(4,0.416)(8,0.446)(16,0.489)(32,0.552)
                };
                \addplot[color=green!50!black, mark=square*, solid] coordinates {
                  (1,0.262)(2,0.344)(4,0.39)(8,0.423)(16,0.456)(32,0.506)
                };
                \addplot[color=red, mark=triangle*, solid] coordinates {
                  (1,0.049)(2,0.095)(4,0.207)(8,0.367)(16,0.517)(32,0.662)
                };
                \addplot[color=blue, mark=triangle*, solid] coordinates {
                  (1,0.028)(2,0.074)(4,0.181)(8,0.301)(16,0.417)(32,0.524)
                };
                \addplot[color=green!50!black, mark=triangle*, solid] coordinates {
                  (1,0.019)(2,0.044)(4,0.119)(8,0.204)(16,0.317)(32,0.418)
                };
                \addplot[color=red, mark=diamond*, solid] coordinates {
                  (1,0.023)(2,0.053)(4,0.118)(8,0.239)(16,0.459)(32,0.709)
                };
                \addplot[color=blue, mark=diamond*, solid] coordinates {
                  (1,0.02)(2,0.046)(4,0.094)(8,0.188)(16,0.348)(32,0.548)
                };
                \addplot[color=green!50!black, mark=diamond*, solid] coordinates {
                  (1,0.013)(2,0.029)(4,0.06)(8,0.121)(16,0.236)(32,0.41)
                };
            \end{axis}
        \end{tikzpicture}
    \end{minipage}}\hfill%

    \vspace{0.3em}
    \ref{legendTopK}\par\vspace{0.2em}
    \ref{legendBatch}

    \caption{Deletion curves showing target drop as a function of number of deleted features (N). Left column: TopKSAE, right column: BatchTopKSAE. Rows: DINO, CLIP and BLIP models, respectively.}
    \label{fig:deletion_curves_target_per_metric}
\end{figure*}
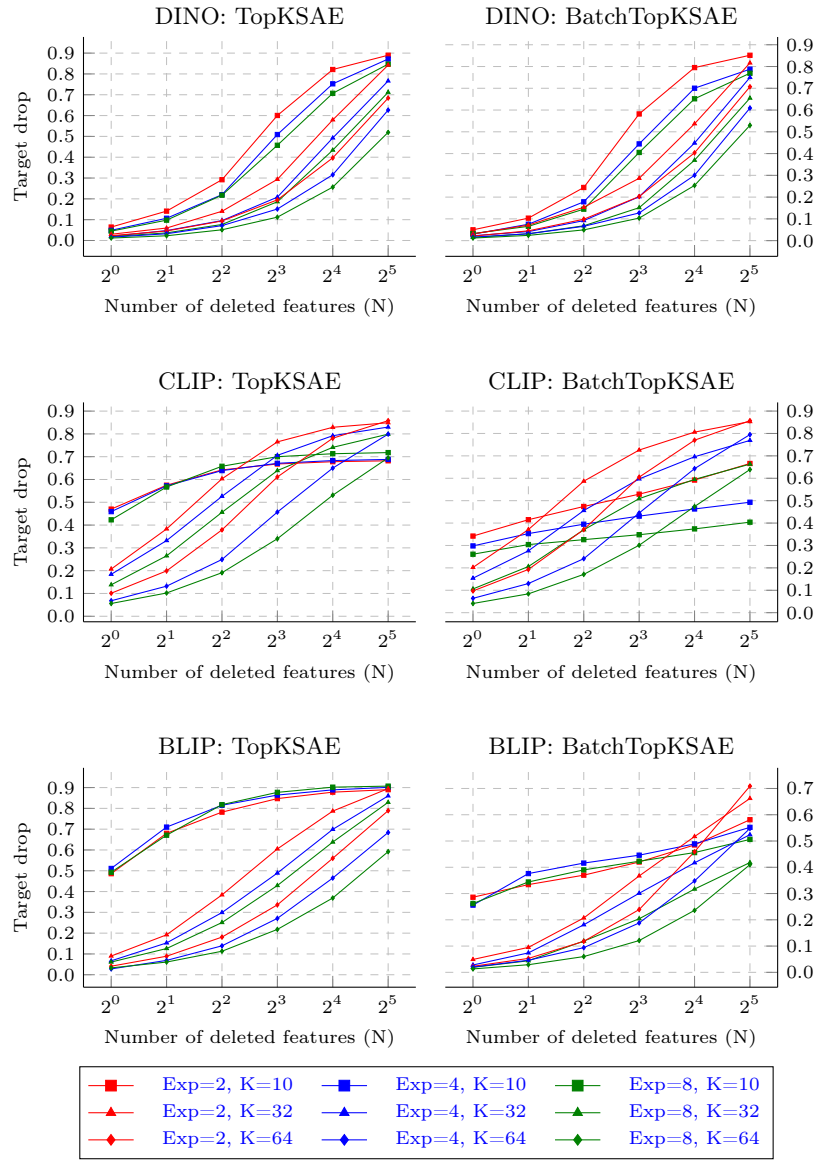

%% file: TABS/stats_with_dead_features_tab.tex
\begin{table*}[t]
    \centering
    \caption{Training metrics of SAE models across different feature extractors and configurations. $R^2$ denotes explained variance, $L_0$ is the fraction of zero activations in latent codes, CS denotes mean cosine similarity between input and reconstruction, and Dead features ratio is the fraction of latent dimensions never activated during training. Best values per feature extractor marked in bold.}
    \label{table:sae_training_metrics_df}
    \begin{tabular}{|c|c|c|cc|cc|cc|cc|}
        \hline
        \multirow{2}{*}[-5pt]{Model} & \multirow{2}{*}[-5pt]{\makecell{Exp.\\Factor}} & \multirow{2}{*}[-5pt]{$K$} & \multicolumn{2}{c|}{$R^2$} & \multicolumn{2}{c|}{$L_0$} & \multicolumn{2}{c|}{CS} & \multicolumn{2}{c|}{\makecell{Dead\\Features}} \\ \cline{4-11}
                               &                               &                      & \makecell{Batch\\TopK} & TopK & \makecell{Batch\\TopK} & TopK & \makecell{Batch\\TopK} & TopK & \makecell{Batch\\TopK} & TopK \\ \hline
        \multirow{9}{*}{DINO} & \multirow{3}{*}{$\times$2} & 10 & 0.67 & 0.68 & 0.99 & 0.99 & 0.80 & 0.81 & 0.001 & 0.000 \\ \cline{3-11}
         &  & 32                                                & 0.71 & 0.71 & 0.97 & 0.98 & 0.84 & 0.83 & 0.000 & 0.000 \\ \cline{3-11}
         &  & 64                                                & 0.72 & 0.72 & 0.95 & 0.96 & 0.85 & 0.84 & 0.000 & 0.000 \\ \cline{2-11}
         & \multirow{3}{*}{$\times$4} & 10                      & 0.69 & 0.70 & 0.99 & 1.00 & 0.81 & 0.83 & 0.003 & 0.000 \\ \cline{3-11}
         &  & 32                                                & 0.73 & 0.73 & 0.99 & 0.99 & 0.85 & 0.84 & 0.000 & 0.000 \\ \cline{3-11}
         &  & 64                                                & 0.75 & 0.74 & 0.98 & 0.98 & 0.86 & 0.85 & 0.000 & 0.000 \\ \cline{2-11}
         & \multirow{3}{*}{$\times$8} & 10                      & 0.70 & 0.72 & 1.00 & 1.00 & 0.82 & 0.83 & 0.036 & 0.003 \\ \cline{3-11}
         &  & 32                                                & 0.76 & 0.75 & 0.99 & 0.99 & 0.86 & 0.85 & 0.000 & 0.000 \\ \cline{3-11}
         &  & 64                                                & 0.77 & 0.76 & 0.99 & 0.99 & 0.86 & 0.86 & 0.000 & 0.000 \\ \hline
        \multirow{9}{*}{CLIP} & \multirow{3}{*}{$\times$2} & 10 & 0.67 & 0.71 & 0.96 & 0.99 & 0.76 & 0.84 & 0.000 & 0.000 \\ \cline{3-11}
         &  & 32                                                & 0.85 & 0.85 & 0.92 & 0.97 & 0.91 & 0.92 & 0.000 & 0.000 \\ \cline{3-11}
         &  & 64                                                & 0.89 & 0.89 & 0.92 & 0.94 & 0.94 & 0.94 & 0.000 & 0.000 \\ \cline{2-11}
         & \multirow{3}{*}{$\times$4} & 10                      & 0.67 & 0.71 & 0.96 & 1.00 & 0.74 & 0.84 & 0.000 & 0.000 \\ \cline{3-11}
         &  & 32                                                & 0.85 & 0.85 & 0.95 & 0.98 & 0.89 & 0.92 & 0.000 & 0.000 \\ \cline{3-11}
         &  & 64                                                & 0.89 & 0.90 & 0.95 & 0.97 & 0.94 & 0.94 & 0.000 & 0.000 \\ \cline{2-11}
         & \multirow{3}{*}{$\times$8} & 10                      & 0.69 & 0.70 & 0.98 & 1.00 & 0.75 & 0.84 & 0.004 & 0.000 \\ \cline{3-11}
         &  & 32                                                & 0.86 & 0.86 & 0.97 & 0.99 & 0.88 & 0.92 & 0.000 & 0.000 \\ \cline{3-11}
         &  & 64                                                & 0.90 & 0.90 & 0.97 & 0.98 & 0.94 & 0.94 & 0.000 & 0.000 \\ \hline
        \multirow{9}{*}{BLIP} & \multirow{3}{*}{$\times$2} & 10 & 0.62 & 0.65 & 0.97 & 1.00 & 0.68 & 0.81 & 0.004 & 0.001 \\ \cline{3-11}
         &  & 32                                                & 0.80 & 0.79 & 0.97 & 0.99 & 0.88 & 0.89 & 0.001 & 0.002 \\ \cline{3-11}
         &  & 64                                                & 0.83 & 0.81 & 0.97 & 0.98 & 0.91 & 0.90 & 0.004 & 0.030 \\ \cline{2-11}
         & \multirow{3}{*}{$\times$4} & 10                      & 0.62 & 0.67 & 0.98 & 1.00 & 0.68 & 0.82 & 0.012 & 0.006 \\ \cline{3-11}
         &  & 32                                                & 0.80 & 0.80 & 0.98 & 0.99 & 0.87 & 0.89 & 0.006 & 0.004 \\ \cline{3-11}
         &  & 64                                                & 0.83 & 0.83 & 0.98 & 0.99 & 0.91 & 0.91 & 0.011 & 0.058 \\ \cline{2-11}
         & \multirow{3}{*}{$\times$8} & 10                      & 0.64 & 0.67 & 0.99 & 1.00 & 0.66 & 0.82 & 0.044 & 0.020 \\ \cline{3-11}
         &  & 32                                                & 0.81 & 0.81 & 0.99 & 1.00 & 0.87 & 0.89 & 0.018 & 0.013 \\ \cline{3-11}
         &  & 64                                                & 0.83 & 0.79 & 0.99 & 0.99 & 0.91 & 0.90 & 0.030 & 0.121 \\ \hline
    \end{tabular}
\end{table*}

%% file: TABS/accuracy_comparison_tab.tex
\begin{table*}[h]
    \centering
    \caption{Reconstruction accuracy of SAE models across different feature extractors and configurations. Baseline Acc denotes linear probe accuracy on raw features. SAE Acc denotes linear probe accuracy on SAE latent representations. Best SAE result per feature extractor marked in bold.}
    \label{table:sae_accuracy}
    \begin{tabular}{|c|c|c|c|c|c|}
        \hline
        \multirow{2}{*}{Model} & \multirow{2}{*}{Exp.\ Factor} & \multirow{2}{*}{$K$} & \multirow{2}{*}{Baseline Acc} & \multicolumn{2}{c|}{SAE Acc} \\ \cline{5-6}
                               &                               &                      &                               & BatchTopK & TopK \\ \hline
        \multirow{9}{*}{DINO} & \multirow{3}{*}{$\times$2} & 10 & \multirow{9}{*}{93.08\%} & 90.90\% & 90.31\% \\ \cline{3-3}\cline{5-6}
         &  & 32 &  & 91.85\% & 91.55\% \\ \cline{3-3}\cline{5-6}
         &  & 64 &  & \textbf{92.36\%} & 92.09\% \\ \cline{2-3}\cline{5-6}
         & \multirow{3}{*}{$\times$4} & 10 &  & 90.49\% & 90.26\% \\ \cline{3-3}\cline{5-6}
         &  & 32 &  & 91.65\% & 91.45\% \\ \cline{3-3}\cline{5-6}
         &  & 64 &  & 92.06\% & 91.99\% \\ \cline{2-3}\cline{5-6}
         & \multirow{3}{*}{$\times$8} & 10 &  & 89.80\% & 90.05\% \\ \cline{3-3}\cline{5-6}
         &  & 32 &  & 91.50\% & 91.54\% \\ \cline{3-3}\cline{5-6}
         &  & 64 &  & 91.82\% & 91.77\% \\ \hline
        \multirow{9}{*}{CLIP} & \multirow{3}{*}{$\times$2} & 10 & \multirow{9}{*}{89.83\%} & 70.93\% & 68.36\% \\ \cline{3-3}\cline{5-6}
         &  & 32 &  & 86.56\% & 85.34\% \\ \cline{3-3}\cline{5-6}
         &  & 64 &  & 87.47\% & 87.25\% \\ \cline{2-3}\cline{5-6}
         & \multirow{3}{*}{$\times$4} & 10 &  & 59.67\% & 68.96\% \\ \cline{3-3}\cline{5-6}
         &  & 32 &  & 86.93\% & 84.87\% \\ \cline{3-3}\cline{5-6}
         &  & 64 &  & 87.87\% & 86.89\% \\ \cline{2-3}\cline{5-6}
         & \multirow{3}{*}{$\times$8} & 10 &  & 56.45\% & 72.06\% \\ \cline{3-3}\cline{5-6}
         &  & 32 &  & 86.67\% & 84.55\% \\ \cline{3-3}\cline{5-6}
         &  & 64 &  & \textbf{88.24\%} & 86.84\% \\ \hline
        \multirow{9}{*}{BLIP} & \multirow{3}{*}{$\times$2} & 10 & \multirow{9}{*}{96.49\%} & 90.14\% & 89.42\% \\ \cline{3-3}\cline{5-6}
         &  & 32 &  & 95.35\% & 94.85\% \\ \cline{3-3}\cline{5-6}
         &  & 64 &  & \textbf{95.47\%} & 95.09\% \\ \cline{2-3}\cline{5-6}
         & \multirow{3}{*}{$\times$4} & 10 &  & 91.07\% & 90.33\% \\ \cline{3-3}\cline{5-6}
         &  & 32 &  & 95.14\% & 94.71\% \\ \cline{3-3}\cline{5-6}
         &  & 64 &  & 95.34\% & 94.89\% \\ \cline{2-3}\cline{5-6}
         & \multirow{3}{*}{$\times$8} & 10 &  & 89.46\% & 90.75\% \\ \cline{3-3}\cline{5-6}
         &  & 32 &  & 94.85\% & 94.55\% \\ \cline{3-3}\cline{5-6}
         &  & 64 &  & 95.09\% & 94.82\% \\ \hline
    \end{tabular}
\end{table*}